\documentclass{article}
\usepackage{jfrExamplee}
\usepackage{graphicx}
\usepackage{apalike}
\usepackage{setspace}
\usepackage{amsmath,bm}
\usepackage{enumerate}
\usepackage{amssymb}
\usepackage[mathscr]{eucal}
\usepackage{relsize}
\usepackage{array}
\usepackage{graphicx}
\usepackage{balance}
\usepackage{hyperref}
\usepackage{array}
\usepackage{url}
\usepackage{caption,subcaption,gensymb}
\usepackage{tabularx}
\usepackage{graphicx}
\usepackage{cite}
\usepackage{graphics} 
\usepackage{algorithm}
\usepackage{algpseudocode}
\usepackage{xcolor}
\usepackage{rotating}
\usepackage{dsfont}

\DeclareMathOperator*{\argmax}{arg\,max}

\newcommand\norm[1]{\lVert#1\rVert}
\newcommand\red[1]{\textcolor{red}{#1}}

\def\BibTeX{{\rm B\kern-.05em{\sc i\kern-.025em b}\kern-.08em
    T\kern-.1667em\lower.7ex\hbox{E}\kern-.125emX}}
\DeclareUnicodeCharacter{2212}{-}
\UseRawInputEncoding

\title{Active Sensing Strategy: Multi-Modal, Multi-Robot Source Localization and Mapping in Real-World Settings with Fixed One-Way Switching}

\author{
Vu Phi Tran \\
University of New South Wales, \\
Canberra, Australia \\
\texttt{phivus2@gmail.com} \\
\And
Asanka G. Perera\\
University of New South Wales, \\
Canberra, Australia\\
\And
Matthew A. Garratt\\
University of New South Wales, \\
Canberra, Australia\\
\And
Kathryn Kasmarik \\
University of New South Wales, \\
Canberra, Australia\\
\And
Sreenatha G. Anavatti \\
University of New South Wales, \\
Canberra, Australia}

\begin{document}

\maketitle

\begin{abstract}
This paper introduces a state-machine model designed for a multi-modal, multi-robot environmental sensing algorithm tailored to dynamic real-world settings. The multi-modal algorithm uniquely combines two distinct exploration strategies for gas source localization and mapping tasks: (1) an initial exploration phase using multi-robot coverage path planning with variable formations, providing early gas field indication; and (2) a subsequent active sensing phase employing multi-robot swarms for precise field estimation. The state machine provides the logic for the transition between these two sensing algorithms. In the exploration phase, a coverage path is generated, maximizing the visited area while measuring gas concentration and estimating the initial gas field at pre-defined sample times. Subsequently, in the active sensing phase, mobile robots moving in a swarm collaborate to select the next measurement point by broadcasting potential positions and reward values, ensuring coordinated and efficient sensing for a multi-robot swarm system. System validation involves hardware-in-the-loop experiments and real-time experiments with a radio source emulating a gas field. The proposed approach is rigorously benchmarked against state-of-the-art single-mode active sensing and gas source localization techniques. The comprehensive evaluation highlights the multi-modal switching approach's capacity to expedite convergence, adeptly navigate obstacles in dynamic environments, and significantly enhance the accuracy of gas source location predictions. \red{These findings highlight the effectiveness of our approach, showing significant improvements: a 43\% reduction in turnaround time, a 50\% increase in estimation accuracy, and enhanced robustness of multi-robot environmental sensing in cluttered scenarios without collisions. These advancements surpass the performance of conventional active sensing strategies, the partial differential equation model, and geometrical localization approaches, underscoring the efficacy of our method.}



\end{abstract}
\section*{Keywords}
Coverage Path Planning, Gas Source Localisation and Mapping, Robotic Flock, Formation Control, Multi-Robot Systems, Multi-Sensor Fusion.


\section{Introduction}
Gas distribution mapping is a crucial task in environmental monitoring, disaster response, and safety inspections, as it enables the accurate estimation of gas concentrations in various contexts \cite{ma2020location, rhodes2020informative, francis2022gas}. However, this task is challenging due to the complexity of the physical mechanisms underlying gas release and the mutual mixture of different plumes, which makes it challenging to obtain precise estimates using static sensors alone \cite{ma2020location,salcedo2020}. To improve estimation performance, mobile devices like robots equipped with various sensors have been proposed \cite{jang2020multi,rosser2021low,nguyen2022characteristics,leong2022field,nguyen2023neural,tran2023multi}.

Despite recent progress, a common issue in gas field mapping with mobile sensors is the selection of measurement locations, which can lead to significant increases in backtracking paths if not selected accurately \cite{wiedemann2019model, jang2020multi, leong2022field, tran2023multi}. This problem is particularly pronounced during initialization, when there is a lack of prior information and high uncertainty in source parameters, resulting in slow convergence to the true gas field \cite{wiedemann2019model, leong2022logistic, tran2023multi}. Furthermore, most multi-robot active sensing operations have been conducted in ideal environments without any obstacles or only static obstacles \cite{fung2019coordinating,hutchinson2019source,leong2021estimation,tran2023multi,leong2024estimation}.

Given that the environment information is partially known—such as the dimensions of the study area and the coordinates of several static obstacles—while the gas distribution and dynamic obstacles remain unknown, an optimal coverage path planning (CPP) algorithm that ensures maximum area coverage and constructs the initial shape of the unknown gas field can significantly reduce estimation time and improve accuracy \cite{huang2020multi,li2020multi,wiedemann2021robotic}.

\red{However, recent multi-robot path planning algorithms, as discussed in \cite{wiedemann2019model, flaspohler2019information, hinsen2023exploration, liao2023experimental, perera2023radio, tran2024collaborative}, primarily focus on localizing the source term by directing multi-robot systems to informative measurement locations with the highest concentration readings. These approaches are unsuitable for gas field estimation in cluttered environments because measurements quickly become concentrated around the source locations, providing more information about the sources but less about the overall field. While decentralized multi-robot exploration with macro actions \cite{flaspohler2019information, tan2022deep, zhang2023heterogeneous} and informative path planning \cite{wiedemann2019model, rhodes2020informative} show potential by combining robotic path planning with source-seeking strategies, obstacle avoidance systems, and gas distribution models, they fail to capture the `qualitative' shape of the true field. These methods rapidly reach the source location without ensuring maximum coverage of the area of interest or taking sufficient measurements around the gas field boundaries, both crucial for accurate gas field estimations.}

\red{In our previous works \cite{tran2023coverage,perera2023radio}, an innovative approach to optimal coverage path planning and obstacle avoidance for multiple Unmanned Ground Vehicles (UGVs) in dynamic environments was introduced, considering constraints on time, path length, number of UGVs, and obstacles. This paper extends the application of the proposed CPP from gas source localization to gas distribution mapping, aiming to provide an initial estimated field for the active sensing process.}

\red{Motivated by research gaps in mapping gas distribution using mobile robots, this paper introduces a novel state machine model for transitioning between CPP and AS, enhancing traditional single-mode mechanisms to achieve faster and more accurate gas field approximation.  It pioneers the use of CPP for initial gas field mapping and proposes a novel collaborative entropy-driven AS approach designed specifically for cluttered environments.} The key contributions are summarized as follows:
\begin{enumerate}
\item \red{Introduction of a novel general state machine model for gas field estimation, detailing the conditions for transitions between the budget-constrained coverage path planning (CPP) algorithm \cite{tran2023coverage,perera2023radio} and the new active sensing (AS) approach. This multi-modal method, which switches between CPP and AS, advances beyond traditional single-mode active sensing mechanisms used in previous studies \cite{leong2021estimation,leong2022field,tran2023multi}.}

\item \red{First application of the CPP algorithm for approximating the initial gas field.}

\item \red{Development of a novel collaborative entropy-driven AS approach, specifically designed for cluttered environments. This approach utilizes LiDAR-based information and leverages collaborative Sequential Monte Carlo (CSMC) techniques with binary measurements and multi-robot swarm behavior \cite{tran2023multi}, effectively managing obstacles and significantly reducing estimation time for autonomous robots.}

\item  A comparative study bench-marking the proposed CPP-AS switching approach against state-of-the-art AS approaches \cite{leong2021estimation,leong2022field,tran2023multi} and source localization methods \cite{wiedemann2019model,guzey2022rf}. The study evaluates various metrics, including estimation accuracy, total exploration time, and swarming metrics, in hardware-in-the-loop and real-time experiments with a radio source simulating a real gas field.

\end{enumerate}

	

\red{In field estimation, instead of assuming specific characteristics for a particular threat using a complex gas field model that may not perform online, the field is often modeled as a sum of radial basis functions (RBFs) \cite{jang2020multi,leong2022field,tran2023multi} or Gaussian mixture models \cite{fung2019coordinating,shi2020adaptive,masaba2023multi}. This approach simplifies the problem to estimating the parameters of the individual basis functions, which, in this paper, is accomplished using sequential Monte Carlo techniques \cite{tran2023multi}. According to approximation theory, these models, especially RBFs, are effective in approximating cross-interference multiple complex fields with arbitrary accuracy as the number of basis functions increases \cite{park1993approximation}. This is the main advantage of RBFs over other complex models like Gaussian Process Regression \cite{jang2020multi,abdul2021scalar}. Additionally, RBF models have demonstrated robustness with binary measurements, further validating their suitability for our purposes \cite{leong2022field,leong2022logistic,tran2023multi}. Thus, while our approach incorporates certain simplifying assumptions, it effectively balances complexity and practical applicability, making it highly relevant for comprehensive gas field estimation.}

\red{For convenience, Tables \ref{tab:variables1} \ref{tab:variables2}, and \ref{tab:variables3} summarise a list of significant parameters used in this manuscript.}

\begin{table}[h]
 \centering
\caption {\red{List of gas model parameters}}
\label{tab:variables1}
  \centering
  \begin{tabular}{ll}
    \hline
     \textbf{Parameter} &  \textbf{Description} \\
      \hline
$\gamma_i$ &  Coefficients of the basis functions \\
$\mu_i$ & Centre of the i$^{th}$ Gaussian basis function  \\
$\sigma_i$ & Width of the i$^{th}$ Gaussian basis function   \\
$I$ & Total number of basis functions   \\
$c(l)$ & Concentration level at the point $l$ \\
$\lambda(l)$ & Sensor noise \\
$\sigma_\delta^2$ & Noise variance \\
$\tau$  & Positive scalar threshold \\
\hline\hline
  \end{tabular}
\end{table}

\begin{table}[h]
 \centering
\caption {\red{List of robotic swarm parameters}}
\label{tab:variables2}
  \centering
  \begin{tabular}{ll}
    \hline
     \textbf{Parameter} &  \textbf{Description} \\
      \hline
($\underline{x_w}$, $\overline{x_w}$) and ($\underline{d_w}$, $\overline{d_w}$) & Positions of the four virtual walls \\
$\Delta l_{rg}$  & Relative position between the robot and the goal \\
$\upsilon_{w(t)}$ & Wall avoidance force vector \\
$\upsilon_{g(t)}$ & Goal attraction force vector \\
$\upsilon_{c(t)}$ & Cohesion force vector \\
$\upsilon_{s(t)}$ & Separation force vector \\
$\upsilon_{a(t)}$ & Alignment force vector \\
$\upsilon_{av(t)}$ & Avoidance force vector \\
$\omega_{c}$ &  Weight for cohesion rule  \\
$\omega_{a}$ &  Weight for alignment rule  \\
$\omega_{s}$ &  Weight for separation rule  \\
$\omega_{w}$ &  Wall Weight \\
$\omega_{at}$ &  Attractive force Weight \\
$R_{a}$ &  Alignment Radius   \\
$R_{s}$ &  Separation Radius   \\
$R_{c}$   &  Cohesion and Communication Radius \\
$R_{av}$ &  Obstacle Avoidance Radius \\
\hline\hline
  \end{tabular}
\end{table}

\begin{table}[h]
 \centering
\caption {\red{List of algorithm parameters}}
\label{tab:variables3}
  \centering
  \begin{tabular}{ll}
    \hline
     \textbf{Parameter} &  \textbf{Description} \\
      \hline

$p(\xi)$ & Probability density for the parameter vector $\xi$ \\
$\mathbb{E}_r[.]$ & Expectation with respect to $r(\xi)$ \\
$\epsilon$ &  Exploration probability   \\
$S_a$ & Neighbouring agents of agent $a$ \\
$m$ & Centre of each Gaussian distribution \\
$\tilde{\bm{\omega}}$ & Weight of each center \\
$D_{d,b}$ & List of neighbor blocks of the block $b$ in the direction $d$ \\
$Q_{c,b}$ & Coordinate of the center of the part $c$ of the block $b$ \\ 
$P_{c,b}$ & Coordinate of the center of the block $b$ \\
$l_o$ & Spring length vector \\
$\Upsilon$ & Radial step size \\
$N_b$ & List of neighbour blocks of the block $b$ \\
$N_l$ &  Number of measurements  \\
$N_p$ &  Number of particles  \\
$N_{s}$ &  Maximum Number of step lengths  \\
$N_d$ & Number of the heading robot directions   \\
$A$ & Allowable search area \\
$F$ & Free space within $A$ \\
$OB$ & Obstacle area within $A$ \\
$\theta$ & Exploration parameter \\
\hline\hline
  \end{tabular}
\end{table}

The remainder of this paper is organized as follows. In Section II we first briefly introduce the system model, formally define the gas distribution mapping problem, and review the background necessary to understand the rest of the paper. Our novel algorithm is presented in Section III. Section IV describes the hardware-in-the-loop, real-radio outdoor, and simulation experiments and results on a 2-dimensional scalar field. In Section V, we conclude the paper.

\section{Background and Related Work}
This section commences with 
an overview of the prior work that closely aligns with the study presented in this paper. Subsequently, the section presents background information that is essential for comprehending the algorithm discussed in the subsequent sections.

\subsection{Related Work}

Field estimation, which includes source localisation and mapping, is a critical task in various applications, such as precision agriculture \cite{oliveira2021advancesgo}, environmental monitoring \cite{nguyen2021mobile}, and search and rescue missions \cite{cao2016distributed}. In recent years, significant research effort has been devoted to developing efficient and accurate algorithms and techniques for field estimation using active sensing \cite{jang2020multi,li2021attention,leong2022logistic}

One widely adopted approach to source localisation and mapping is based on probabilistic models, such as Bayesian inference and Kalman filtering \cite{pourbabaee2015sensor,xue2017bayesian}. These methods use multiple sensors deployed at fixed locations to estimate the location and properties of a signal source. However, their accuracy is limited by the number and arrangement of the sensors, and they assume a known sensor model and Gaussian noise, which may not hold in many real-world scenarios.

Machine learning techniques, such as neural networks and support vector machines, have also been applied to field estimation \cite{salcedo2020}. For instance, in \cite{tao2019air,kow2022real}, a deep learning-based method was used to predict air pollutant concentrations in urban areas using a wireless sensor network. Similarly, in a study by \cite{yuan2020estimating}, a neural network-based method was developed for estimating soil moisture levels using data from satellite and ground sensors. However, these methods require large amounts of training data, which may be difficult to obtain in certain scenarios.

Active sensing, employing a gas sensor on each mobile robotic platform, such as an unmanned ground vehicle (UGV) or unmanned aerial vehicle (UAV), actively guides the exploration process and has shown significant promise in field estimation \cite{carrozzo2018uav,ji2021multi,ercolani2022clustering,tran2023multi}. Unlike traditional passive sensing, active sensing dynamically controls sensor-mounted robots to optimize information gain while minimizing costs \cite{nelson2006sensory}.

In single-robot systems, information-theoretic measures like entropy or mutual information can guide the robot's actions towards the most uncertain areas. Algorithms such as Active Search \cite{asenov2019active}, Information-Guided Exploration \cite{flaspohler2019information}, and Adaptive Sampling \cite{arain2021sniffing} have been developed for this purpose.

\red{For multi-robot systems, active sensing becomes more effective as robots collaborate to reduce estimation uncertainty and cover larger areas \cite{ma2020location,tran2023multi}. Active sensing can be performed using individual robot behaviors, where each robot uses its own sensor data to build a map of gas concentrations \cite{wiedemann2017bayesian,leong2022field,shi2020adaptive,masaba2023multi}. However, strategies like adaptive sampling with static or adaptive Voronoi partitioning \cite{salam2019adaptive,shi2020adaptive,ercolani2023multi,masaba2023multi} encounter challenges. These include increased computational costs for path planning and fused gas field estimation as the number of robots increases. Additionally, these methods are limited to estimating a single gas field in open spaces due to the absence of obstacle avoidance mechanisms and gas field mixture models. Furthermore, the failure of a single robot can lead to incomplete mapping.}

\red{An alternative solution, such as Gaussian process regression (GPR) \cite{jang2020multi,abdul2021scalar}, offers a distributed multi-robot exploration approach to enhance system robustness. However, this method assumes real-valued (non-quantized) sensor measurements, which can lead to noisy readings that are coarsely quantized to a few levels (up to 4 or 5). Additionally, GP regression is less effective in scenarios requiring binary measurements, where sensors only detect the presence or absence of chemicals \cite{rosser2015autonomous}.}

\red{The use of mobile agents with binary measurements for field estimation is discussed in \cite{leong2021estimation,leong2022field,leong2022logistic}. However, these methods are less effective in large-scale settings due to limited communication ranges. Additionally, the independent sensing actions of each robot often result in inconsistent gas field maps, prolonged completion times, and low-accuracy approximations, particularly in expansive areas, as demonstrated in Experiments 1-2 of \cite{tran2023multi}.}

\red{Formation control \cite{wang2017leader,tran2020distributed,tran2020switching,tran2020switching1,wang2022nonlinear},  another multi-robot cooperative strategy, typically requires agents to maintain specific geometric configurations, which limits their adaptability in dynamic and unpredictable environments. As the number of agents increases, the complexity of maintaining a stable formation also grows, making it computationally intensive and challenging. This approach can be less efficient for exploring large or complex areas, as the need to maintain formation restricts the agents' ability to cover more spaces quickly and take measurements from diverse locations.}

\red{Moreover, if the formation control strategy relies on a leader-follower model, the failure of the leader can disrupt the entire formation, making the system vulnerable to single points of failure. In cluttered or highly dynamic environments, maintaining formation can be particularly challenging. Agents may struggle to navigate tight spaces or respond effectively to rapidly changing conditions.}

\red{A more flexible alternative is distributed swarming, which offers adaptability and redundancy, mitigating the impact of individual robot failures. This approach enhances the system's resilience and effectiveness in diverse and unpredictable environments.}

To overcome these limitations, collaborative active sensing can be employed using a swarm, such as flocking or swarming behaviors \cite{li2020multi,tran2023multi}. In this approach, the active sensing algorithm is modified for a robotic flocking system, where all robots select and move towards the same measurement locations in a distributed manner. This approach can reduce exploration time and improve estimation accuracy.

While collaborative active sensing through distributed swarms holds promise, critical challenges must be addressed for real-time applications or emergencies. As discussed in Section I, the primary impediment is the time-intensive nature of mapping with active-sensing techniques, particularly with poorly chosen starting points. The assumption of obstacle-free or static-obstacle areas during sensing is impractical in real-world environments, complicating the process and potentially selecting measurement positions within obstacles.

Addressing these challenges requires incorporating obstacle avoidance mechanisms within active sensing and swarm algorithms and implementing an active sensing switching strategy. This strategic fusion is vital for enhancing the robustness and efficiency of collaborative active sensing in dynamic environments, paving the way for broader adoption in practical, time-sensitive scenarios.

The next section introduces several fundamental settings to address these issues and discusses a Gaussian plume model used for gas field estimation.

\subsection{Gas Plume Model and Problem Statement}\label{sub:GPM}
\red{The most significant advantage of Gaussian Plume models is that they have a speedy and almost immediate response time and low computational cost, enabling their application in real-time. Moreover, it is possible to implement Gaussian models for more than one source and combine them through superposition}. The concentration $c(l)$ of the field at any point $l(x,y) \in \mathbb{R}^2$ is expressed by the following equation:

\begin{equation}\label{eq:Gauss_field}
c(l) = \sum_{i=1}^{I}\gamma_i~exp\big(\frac{-||\mu_i-l_i||^2}{\sigma_i}\big),
\end{equation}
where $\mu_i \in \mathbb{R}^2$ and $\sigma_i \in \mathbb{R}$ represent the centre and the width of the i$^{th}$ Gaussian basis function, respectively. $\gamma_i$ are the coefficients of the basis functions. Additionally, $I$ is the total number of basis functions that approximate the gas field. 

We assume a measurement model at position $p$, including a noisy measurement:
\begin{equation}
y(l) = c(l) + \lambda(l),
\end{equation}
where $\lambda(l) \in \mathcal{N}(0,\sigma_\delta^2)$ is the sensor noise, modelled as Gaussian signal. Depending on the type of contaminant and the sensor technology, the noise variance $\sigma_\delta^2$ is unknown. This noisy reading is then converted into a binary measurement through a threshold value:
\begin{equation}\label{eq:binary_measurements}
r(l) = \mathds{1}\Big(y(l) >  \tau \Big),
\end{equation}
where $\tau$ is a known positive scalar threshold, and $\mathds{1}(.)$ is the indicator function. 

\red{In chemical attack scenarios, accurately determining the concentration of chemicals poses significant challenges due to their low concentrations. Present-day chemical sensors face limitations caused by internal noise and variability, often yielding readings with a limited number of levels or solely indicating the presence or absence of the chemical, resulting in binary measurements \cite{leong2022field}. These binary measurements, which encode only 1-bit of information, provide the advantage of minimal communication requirements, particularly valuable in wireless communication scenarios \cite{ristic2016localisation}.}

\begin{figure}[ht!]
    \centering 
    \includegraphics[width=2.3in]{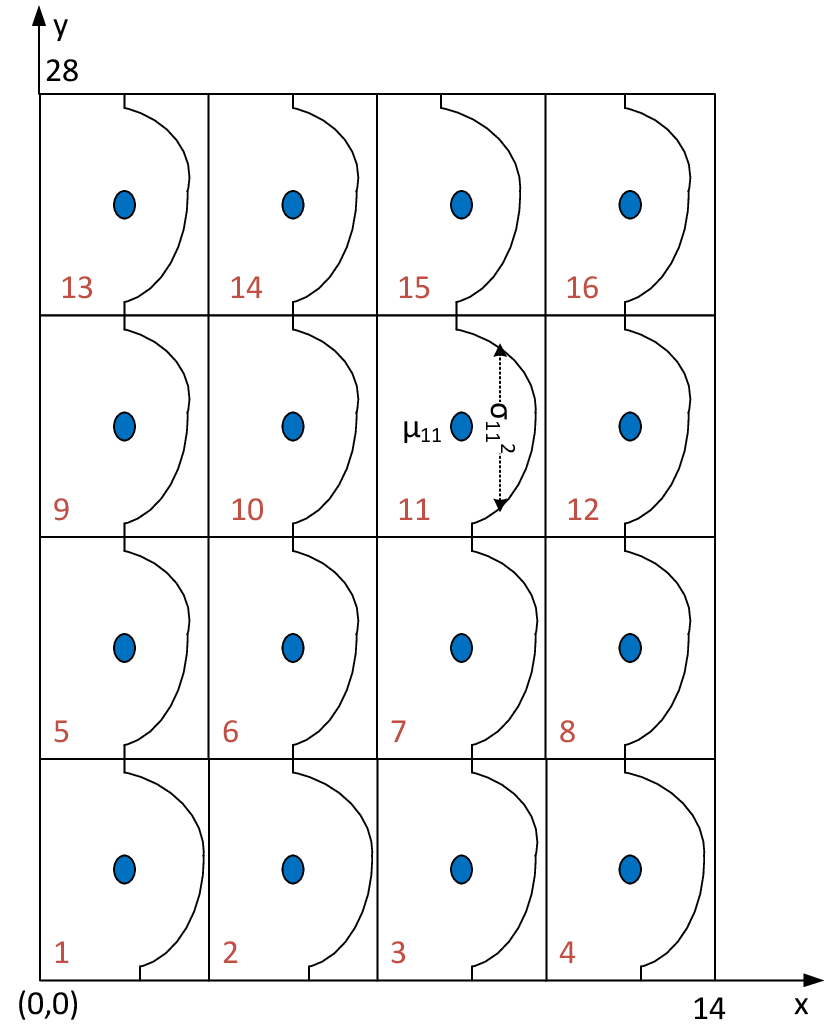}
    \caption{Even Distribution of 16 Gaussian basis functions on the gas mapping area.}
    \label{fig:bf_dis}
\end{figure}

\red{The work described in this paper aims to predict the field (via estimation of the field parameters) from binary measurements without prior knowledge of the actual parameters. The parameters $\mu_i$ and $\sigma_i^2$, $i = 1,..., I$, are systematically distributed to cover the entire map (see Fig. \ref{fig:bf_dis}), while the parameters $\gamma \triangleq (\gamma_1,...., \gamma_{I})$ are estimated from binary measurements collected by a number of agents in a bounded area of interest $\Omega \in \mathbb{R}^2$ while maintaining the various formations.} Each agent estimates gas and chemical dispersion across a partially known area (with available information about static obstacles and the boundary of the exploration area) and shares estimated data with neighbours. The robot determines its following location to travel based on a switching exploration strategy. In the initial phase, the coverage path planner generates comprehensive and safe trajectories, overlaying the simulated concentration field. Following them, the multi-robot system is able to measure gas concentrations at a pre-defined frequency and predict the unknown parameters of the estimated Gaussian field. After the robots return to the starting point, the first exploration strategy is switched to an active sensing mechanism, where a search for the maximum uncertainty position based on the present gas map is prioritized. We consider the following setting:

\begin{itemize}
    \item While the exploration area is dynamic and partially known, the gas field \red{can be adequately approximated by a sum of time-invariant, continuous, and smooth Gaussians}.
    \item Each agent can communicate with other agents within the communication range. Communication between any two agents is bidirectional.
    \item All measurements and estimates are performed online.   
\end{itemize}

To compute an expectation, $\pi_f = \mathbb{E}_p[f(\xi)]$, where $p(\xi)$ is the probability density for the parameter vector $\xi$, we introduce a probability density $r(\xi)$ satisfying $r(\xi) > 0$ whenever $f(\xi)~p(\xi) \neq 0$. This leads to:
\begin{equation}
\pi_f = \int f(\xi)~p(\xi)~d\xi. 
\end{equation}

\subsection{\red{Sequential Monte Carlo method and convergence results}}\label{subsec:convergence}

SMC methods, commonly referred to as particle filters, are a computational approach employed in Bayesian filtering for real-time estimation of dynamic systems. It is particularly useful when dealing with non-linear and non-Gaussian state space models. The method involves representing the probability distribution of a system's state using a set of discrete samples, or particles, which are sequentially updated as new observations become available. The algorithm consists of prediction and update steps, where particles are propagated forward based on the system's dynamic model and then weighted according to their likelihood given the observed data. Resampling is performed to maintain a representative set of particles, providing an efficient means of approximating the evolving posterior distribution of the system's state over time \cite{alex2}. 

To compute an expectation $\pi_f = \mathbb{E}_p[f(\xi)]$, with $p(\xi)$ representing the probability density for the parameter vector $\xi$, we consider a probability density $r(\xi)$ satisfying $r(\xi) > 0$ whenever $f(\xi)~p(\xi) \neq 0$. This leads to the expression:
\begin{equation}
\pi_f = \mathbb{E}_r[\omega(\xi)~f(\xi)],
\end{equation}
where $\omega(\xi) = \frac{p(\xi)}{r(\xi)}$, and now $\mathbb{E}_r[.]$ denotes the expectation with respect to $r(\xi)$.

Consequently, a sample of independent draws $\xi^{(1)}, \ldots, \xi^{(m)}$ from $r(\xi)$ can be leveraged to estimate $\pi_f$, with:
\begin{equation}
\hat{\pi}_f = \frac{1}{m} \sum_{i=1}^{m} \omega(\xi^{(i)})~f(\xi^{(i)}).
\end{equation}

In scenarios where the density $p(\xi)$ is known only up to a normalizing constant, such that $\omega (\xi) = c~\omega_0 (\xi)$, where $\omega_0 (\pi)$ can be computed exactly but the multiplicative constant $c$ remains unknown, one employs the ratio estimate to replace $\hat{\pi}_f$:
\begin{equation}
\tilde{\pi}f = \frac{\sum_{i=1}^{m} \omega(\xi^{(i)})~f(\xi^{(i)})}{\sum_{i=1}^{m} \omega(\xi^{(i)})}.
\end{equation}


The convergence of both $\hat{\pi}_f$ and $\tilde{\pi}_f$ to $\pi_f$ as $m \rightarrow \infty$ is established through the strong law of large numbers, a foundational principle in Monte Carlo integration \cite{geweke1989bayesian}.  Furthermore, the central limit theorem, meticulously detailed in \cite{chopin2004central}, affirms that $\sqrt{m} (\hat{\pi}_f - \pi_f)$ and $\sqrt{m} (\tilde{\pi}_f - \pi_f)$ asymptotically follow a normal distribution with mean zero. Their respective variances are characterized by $\mathbb{E}_r\Big[\big(\omega(\xi)~f(\xi) - \pi_f\big)^2\Big]$ and $\mathbb{E}_r\Big[\omega(\xi)^2~\big(f(\xi) - \pi_f\big)^2\Big]$, under the condition that these quantities remain finite.

To consistently estimate these asymptotic variances, the sampled $\xi^{(i)}$ values can be effectively repurposed. The estimators take the form of $\frac{1}{m} \sum_{i} \Big[\omega(\xi^{(i)})~f(\xi^{(i)}) - \hat{\pi}_f)\Big]^2$ and $\sum_{i} \Big[\omega(\xi^{(i)})^2~(f(\xi^{(i)}) - \tilde{\pi}_f)^2 \Big] / \Big[\sum_{i} \omega(\xi^{(i)})\Big]^2$. These estimators provide reliable assessments of the accuracy of the respective approximations, $\hat{\pi}_f$ and $\tilde{\pi}_f$, as the sample size grows, thereby offering valuable insights into the stability and precision of the estimation process.

\subsection{Flocking behavior} \label{sub:FB}
The boids model is a program initially developed by Craig Reynolds in 1986, which replicates the flocking behaviour of birds \cite{reynolds1987}. Each boid steers itself based on a straightforward set of rules. These rules are summarized as follows:

\begin{itemize}
    \item \textbf{Alignment:} Steer towards the average heading of neighbours.
    \item \textbf{Cohesion:} Steer and move towards the average position of neighbours.
    \item \textbf{Separation:} Steer to avoid hitting neighbours.
    
\end{itemize}

Further, three additional rules, namely obstacle avoidance, goal attraction, and virtual wall, are included to navigate a cluttered environment without collision and maintain the robot's flocking behaviour within a specific space. Firstly, a virtual wall encloses the virtual search area. If any robot moves out of the boundaries, a wall avoidance force vector of the $j^th$ robot at the time step $t$, $\upsilon_{w(t)}^{(k)} \in \mathbb{R}^2$, will be performed to steer this robot back into the area of interest:
\begin{equation}
\upsilon_{w(t)}^{(k)}= \omega_w (\upsilon_{wx(t)}^{(k)} + \upsilon_{wy(t)}^{(k)}),
\end{equation}
where,
\begin{equation}
\upsilon^{(k)}_{wx(t)}=\begin{cases}
~(1, 0)~~~~~~~if~{x(t)}^{(k)} < \underline{x_w}\\
~(0, 0)~~~~~~~if~\underline{x_w} \leq	{x(t)}^{(k)} \leq \overline{x_w}\\
(-1, 0)~~~~~~if~{x(t)}^{(k)} > \overline{x_w},
\end{cases}                                      
\end{equation}
\begin{equation}
\upsilon^{(k)}_{wy(t)}=\begin{cases}
(0, 1)~~~~~~~~if~y(t)^{(k)}<\underline{d_w}\\
(0, 0)~~~~~~~~if~\underline{d_w}\leq	{y(t)}^{(k)}\leq \overline{d_w}\\
(0, -1)~~~~~~if~y(t)^{(k)}>\overline{d_w},
\end{cases}                                      
\end{equation}
here, ($\underline{x_w}$, $\overline{x_w}$) and ($\underline{d_w}$, $\overline{d_w}$) indicate the positions of the four virtual walls. Additionally, a virtual wall weight $\omega_w$ determines how quickly the robots drive back to the area of interest.

In addition to the virtual wall vector, \red{a goal attraction vector is individually computed, proportional to the relative position of each robot with respect to the goal}:
\begin{equation}
\upsilon_{g(t)}^{(k)}= \Delta l_{rg},
\end{equation}
where $\Delta l_{rg} \in  \mathbb{R}^2$ is the relative position between the robot and the goal.

Similar to the force calculation formula for environment boundaries, to improve system safety, an intelligent obstacle detection and obstacle avoidance system using a 2D Light Detection and Ranging (LiDAR) sensor (SICK LMS111) is used to detect mobile and static obstacles and inform the control system to stop or go around when travelling to the destination point in an uncertain or unknown environment. The LiDAR data collection module constantly collects the distance-angle pair of the LiDAR beams as the LiDAR sensor releases the laser beams. 
If the robot detects any obstacle in its field of view, these laser lines will be split into two sets of angles: safe and blocked. The safe angles areas are scanned to seek the closest safe angle with respect to the robot's LiDAR sensor mounting position and heading direction. After the closest safe angle is found, the obstacle avoidance strategy steers the robot towards it to evade the obstacle. More explanations can be seen in our previous work \cite{tran2023dynamic}.

Including those vectors, the final resultant force vector $\upsilon_{f(t+1)}^i$ acting on the particular agent is the summation of all component forces with corresponding weights counted for each rule:
\begin{equation}\label{eq:fused_force}
\begin{split}
\upsilon_{f(t+1)}^k =& \upsilon_{f(t)}^k + \omega_c~\upsilon_{c(t)}^k + \omega_s~\upsilon_{s(t)}^k + \omega_a~\upsilon_{a(t)}^k \\
& + \omega_w \upsilon_{w(t)}^k + \omega_g \upsilon_{g(t)}^k + \omega_{av} \upsilon_{av(t)}^k,
\end{split}
\end{equation}
where the behavior model includes key force vectors, including cohesion ($\upsilon_{c(t)}$), separation ($\upsilon_{s(t)}$), alignment ($\upsilon_{a(t)}$), and obstacle avoidance ($\upsilon_{av(t)}$). Each of these force vectors is carefully weighted using positive weights, namely \(\omega_c\), \(\omega_a\), \(\omega_s\), \(\omega_w\), \(\omega_g\), and \(\omega_{av}\), to finely tune the robot's behavior. \red{The values of these parameters were manually configured and fine-tuned using established rules of thumb for Boid swarms \cite{reynolds1987} and guided by insights gained from our previous work with swarming robots \cite{tran2022frontier}.}

Additionally, we introduce the concept of the swarming force vector $\upsilon_{f(t)}$, which is the cumulative sum of the aforementioned force vectors. By accumulating the forces over time, the swarming force vector contributes to the stabilization of robot motion, enabling smoother and more controlled trajectories. Without this force accumulation mechanism, the robot's motion direction would exhibit continuous fluctuations, ultimately leading to system instability.

Finally, the $k^{th}$ robot's linear and angular velocity reference commands, named $V$ and $\omega$ respectively, can be derived from the $x$ and $y$ components of the final force vector given in Eq. \ref{eq:fused_force} using:
\begin{equation}
\begin{split}
V^k &= \sqrt{(\upsilon^k_{x})^2 + (\upsilon^k_{y})^2} \\
\omega^k &= atan2(\upsilon_{y}^k, \upsilon_{x}^k).
\end{split}
\end{equation}


\subsection{Collaborative particle filter-based gas field estimation}

If all parameters $\sigma_\delta^2$, $\gamma_i$, $\mu_i$, $\mu_i^2$, where $i = 1,..., I$, are known, the probabilities of receiving a $0$ or $1$ binary reading at location $l$ will be:

\begin{equation}
\begin{split}
\mathbb{P}(r(l) = 0) &= \mathbb{P}(y(l) <  \tau) \\
&= \mathbb{P}\Big(\lambda(l) < \tau - \sum_{i=1}^{I}\gamma_i~exp\big(\frac{-||\mu_i-l_i||^2}{\sigma_i}\big)\Big) \\
&= \Delta\bigg(\frac{1}{\sigma_\delta}\Big(\tau - \sum_{i=1}^{I}\gamma_i~exp\big(\frac{-||\mu_i-l_i||^2}{\sigma_i}\big)\Big)\bigg),
\end{split}
\end{equation}

\begin{equation}
\mathbb{P}(r(l) = 1) = 1 - \mathbb{P}(r(l) = 0),
\end{equation}
where $\Delta(.) \triangleq \int_{-\infty}^{x} \frac{1}{\sqrt{2\pi}}\exp\big(\frac{-t^2}{2}\big)dt$ denotes the cumulative distribution function of the normal distribution $\mathcal{N}(0,1)$.

\red{As mentioned in Section \ref{sub:GPM}, we consider the case where $\mu_i$, $\sigma_i^2$, $i = 1,..., I$, have been chosen at certain values. However, the unknown $\gamma_i$, $i = 1,..., I$, and $\sigma_\delta^2$ are estimated}. Define a parameter vector:
\begin{equation}
\xi \triangleq  (\gamma_i, log~\sigma_\delta).
\end{equation}

The posterior Probability Distribution Function (PDF) $p(\xi|y;x)$ is approximately sampled using a set of particles $\xi^{(i)}, i = 1,...,N_p$, and relevant weights $\omega^{(i)}, i = 1,..., N_p$, when the mobile robot moves to separate locations and carries out measurements.

The conditional mean estimates for the $I$ number of basis functions can be derived as:
\begin{equation}\label{eq:alpha_hat}
\hat{\gamma}_{i}=\sum_{i=1}^{N_p}\omega^{(i)}\xi_{j}^{(i)},~\forall j = 1,..., I
\end{equation}

\begin{equation}\label{eq:sigma_v}
\hat{\sigma}_{\delta}^{2}=\sum_{i=1}^{N_p}\omega^{(i)}\exp(2\xi_{I+1}^{(i)}),
\end{equation}
where $\xi^{(i)} \triangleq (\xi_{j}^{(i)},..., \xi_{I+1}^{(i)})$.

To ensure consistent local gas maps produced by mobile robots in a collaborative sensing system, a collaborative particle filter method that facilitates the sharing of particles and weights among agents is designed. Whenever an agent acquires a new measurement, it promptly shares its updated particles and weights with its neighboring agents. This collaborative sharing results in a fused list of particles that incorporate the collective knowledge of the entire agent network.

Let $S_{a}$ denote the set of neighbouring agents of agent $a$, including agent $a$ itself. The particles (the 16 basis function gains and a sensor noise) and weights of agent $b \in S_{a}$ are represented as $\{\bm{\xi}_{b}^{(i)}\}$ and $\{ \bm{\omega}_{b}^{(i)} \}$, respectively. To ensure that the sum of weights for all shared and individual particles is equal to 1 (see lines 7-8 of Algorithm \ref{alg:particle_sharing}), we normalize them accordingly, given that $\sum_{b\in{S}_{a}} \sum_{i=1}^{N_p} \bm{\omega}_{b}^{(i)} = |S_{a}|$.


 In accordance with Line 11 of Algorithm \ref{alg:particle_sharing}, the centre of each Gaussian distribution $\textbf{m}_{b,t_b-1}^{(j)}$ is computed for each particle. This computation involves shifting the center of all particles. Subsequently, the weight $\tilde{\bm{\omega}}{b,t_b}^{(j)}$ associated with each center $\textbf{m}_{b,t_b-1}^{(j)}$ is re-weighted using the likelihood function, as indicated in Line 12 of Algorithm \ref{alg:particle_sharing}: 
\begin{equation}
\tilde{\bm{\omega}}_{b,t_b}^{(j)} \propto  p(y_{a,t_a}|\textbf{m}_{b,t_b-1}^{(j)}; \mathbf{x}_{a,t_a}) \bm{\omega}_{b,t_b-1}^{(j)}
\end{equation}

In Algorithm \ref{alg:particle_sharing}, lines 16-18 represent a critical step involving $N_p$ sampling iterations from a set of $N_p \times |S|$ candidates. The selection probability of each candidate is determined by its calculated weight, where higher weights correspond to higher selection probabilities. This sampling process considers candidates from the agent's own set as well as those from neighboring agents. Subsequently, in order to maintain the same number of particles within each agent, the number of candidates is reduced to $N_p$. The resulting candidates are then resampled using a normal distribution, with the mean centered at $ \textbf{m}_{b,t_b-1}^{(j)}$ and the variance calculated as $h^{2-\eta} \textbf{V}_{a,t_a-1}$. Once the new particles are obtained, their weights undergo re-evaluation as part of the subsequent stage:
\begin{equation}
\bm{\omega}_{a,t_a}^{(j)} \propto \frac{p(y_{a,t_a}|\bm{\xi}_{a,t_a}^{(j)}; \mathbf{x}_{a,t_a})}{p(y_{a,t_a}|\textbf{m}_{b^-,t_{b^-}-1}^{(j^-)}; \mathbf{x}_{a,t_a})}
\end{equation}

\red{Furthermore, with the incorporation of resampling, as previously demonstrated in Section \ref{subsec:convergence}, the expected basis function gains and sensor noise, $\hat{\gamma}{i}$ and $\hat{\sigma}_{\delta}^{2}$, exhibit convergence to the true ones, $\gamma_{i}$ and $\sigma_{\delta}^{2}$ as $N_p \rightarrow \infty$, with variances of $h^{2-\eta} \textbf{V}_{a,t_a-1}$.}

Using a set of $\xi$ and $\omega$ values derived from Algorithm \ref{alg:particle_sharing}, $\hat{\gamma}{i}$ and $\hat{\sigma}{\delta}^{2}$ are computed using (\ref{eq:alpha_hat})-(\ref{eq:sigma_v}). These values, in conjunction with $\mu_i$ and $\sigma_i^2$, enable the approximation of the gas diffusion field through the summation of all Gaussian functions.

\begin{algorithm}
\caption{Collaborative particle fusion approach at agent $a$ \cite{tran2023multi}}
\label{alg:particle_sharing}
\begin{algorithmic}[1]
\State \textbf{Algorithm Parameters}: $N_p \in \mathbb{N}$, $c \in (0,1)$,  $h = \sqrt{1-s^2}$, $\eta \geq 0$, prior PDF $p_{0}(\bm{\xi})$, step size $t$
\State \textbf{Inputs}: Initial location $\mathbf{x}_{a,1}$
\State \textbf{Outputs}: Particles $\{\bm{\xi}_{a,t_a}^{(j)}\}$ and weights $\{ \bm{\omega}_{a,t_a}^{(j)} \}$
\State Sample particles $\bm{\xi}_{a,0}^{(j)}, j=1,\dots,N_p$ from $p_{0}(\bm{\xi})$, and assign weights $ \bm{\omega}_{a,0}^{(j)} = \frac{1}{N_p}, j=1,\dots,N_p$
\For{$t_a=1,2,\dots,$}
        \State Observe $y_{a,t_a}$ at location $\mathbf{x}_{a,t_a}$
        \State Set ${S}_{a,t_a}$ = neighbour set of agent {$a$} including agent {$a$}
        \State Normalize $\{{\bm{\omega}}_{b,t_b}^{(j)}\}$ such that $\sum_{b\in{S}_{a,t_a}} \sum_{j=1}^N {\bm{\omega}}_{b,t_b}^{(j)} = 1$
        \For{${b} \in {S}_{a,t_a}$}
                \For{$j=1,\dots,N$} 
                        \State Compute $ \textbf{m}_{b,t_b-1}^{(j)} = s \bm{\xi}_{b,t_b-1}^{(j)}  + (1-s) \bm\bar{\bm{\xi}}_{a,t_a-1}$, where $ \bar{\bm{\xi}}_{a,t_a-1}  = \sum_{b\in{S}_{a,t_a}} \sum_{j=1}^N \bm{\omega}_{b,t_b-1}^{(j)} \bm{\xi}_{b,t_b-1}^{(j)}$  
                        \State Assign \begin{align*}\tilde{\bm{\omega}}_{b,t_b}^{(j)} &\propto  p(y_{a,t_a}|\textbf{m}_{b,t_b-1}^{(j)}; \mathbf{x}_{a,t_a}) \bm{\omega}_{b,t_b-1}^{(j)}\end{align*} 
                \EndFor
        \EndFor
        \State Normalize $\{\tilde{\bm{\omega}}_{b,t_b}^{(j)}\}$ such that $\sum_{b\in{S}_{a,t_a}} \sum_{j=1}^N \tilde{\bm{\omega}}_{b,t_b}^{(j)} = 1$
        \State Sample $N_p$ times with replacement a set of indices $\{j^-: j=1,\dots,N;b^-: b \in S_{a,t_a}\} $, from a distribution with probabilities $\mathbb{P}(j^- = i,b^- = b) = \tilde{\bm{\omega}}_{b,t_b}^{(i)}$
        \For{$j=1,\dots,N$}
                \State Sample a particle $\bm{\xi}_{a,t_a}^{(j)} \sim \mathcal{N}(\textbf{m}_{b^-,t_{b^-}-1}^{(j^-)}, h^{2-\eta} \textbf{V}_{a,t_a-1})$, where $\textbf{V}_{a,t_a-1}  = \sum_{j=1}^N \bm{\omega}_{a,t_a-1}^{(j)} (\bm{\xi}_{a,t_a-1}^{(j)} - \bar{\bm{\xi}}_{a,t_a-1}) (\bm{\xi}_{a,n_a-1}^{(j)} - \bar{\bm{\xi}}_{a,t_a-1})^T$ 
                \State Assign weights $ \bm{\omega}_{a,t_a}^{(j)} \propto \frac{p(y_{a,t_a}|\bm{\xi}_{a,t_a}^{(j)}; \mathbf{x}_{a,t_a})}{p(y_{a,t_a}|\textbf{m}_{b^-,t_{b^-}-1}^{(j^-)}; \mathbf{x}_{a,t_a})} $ \label{line:measurement_fusion_likelihood2}    
        \EndFor
        \State Normalize $\{ \bm{\omega}_{a,t_a}^{(j)} \}$ such that $\sum_{j=1}^N \bm{\omega}_{a,t_a}^{(j)} = 1$
        \State Broadcast $\{ \bm{\xi}_{a,t_a}^{(j)} \}$,$\{ \bm{\omega}_{a,t_a}^{(j)} \}$ to neighbouring agents
        \State Determine  $\mathbf{x}_{a,t_a+1} = \texttt{Active Sensing}(\mathbf{x}_{a,t_a}, \{\bm{\xi}_{a,t_a}^{(j)}\})$ using Algorithm~\ref{alg:c_active_Sensing}   
        \State Receive transmissions from other agents while travelling to $\mathbf{x}_{a,t_a+1}$, and extract new information $\mathcal{SP}_{{S}_{a,t_a+1}} = \{(\bm{\xi}_{b,t_b}^{(j)},\bm{\omega}_{b,t_b}^{(j)}) \} $ 
\EndFor
\end{algorithmic}
\end{algorithm}

\section{Combining multi-robot coverage and active sensing algorithms}

This section introduces a state-machine model that combines two distinct exploration algorithms: coverage path planning using flexible formations and collaborative active sensing using multi-robot swarms. The integration of these algorithms aims to enhance efficiency, reduce exploration time, and improve the overall performance of environmental sensing tasks. The section is structured into three parts.  First, Section \ref{sub:PSA} outlines a general model for switching between coverage path planning and active sensing. It also presents the variant of this model studied in this paper. Next, Section \ref{sub:MCPP} details our coverage path planning approach, while Section \ref{sub:MRSLAS} describes our multi-robot active sensing algorithm.

\subsection{Switching Strategy for Multi-modal Exploration}\label{sub:PSA}

\begin{figure}
	\centering
	\begin{subfigure}[b]{0.495\textwidth}
		\includegraphics[width=\textwidth]{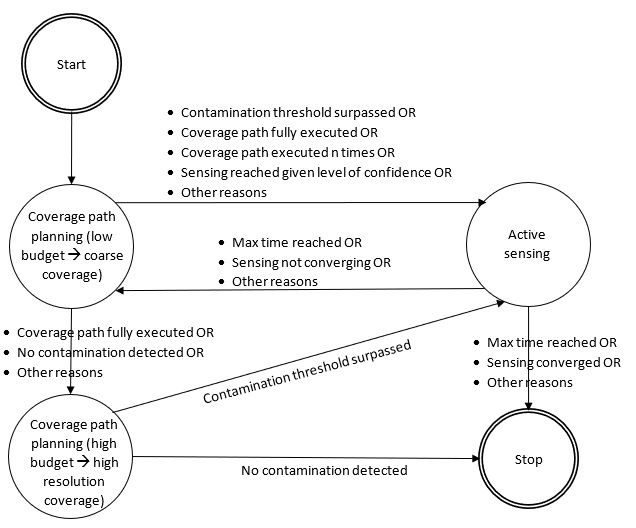}
		\caption{\textit{General multi-modal switching model.}}
		\label{fig:full_state}
	\end{subfigure}
	\begin{subfigure}[b]{0.29\textwidth}
		\includegraphics[width=\textwidth]{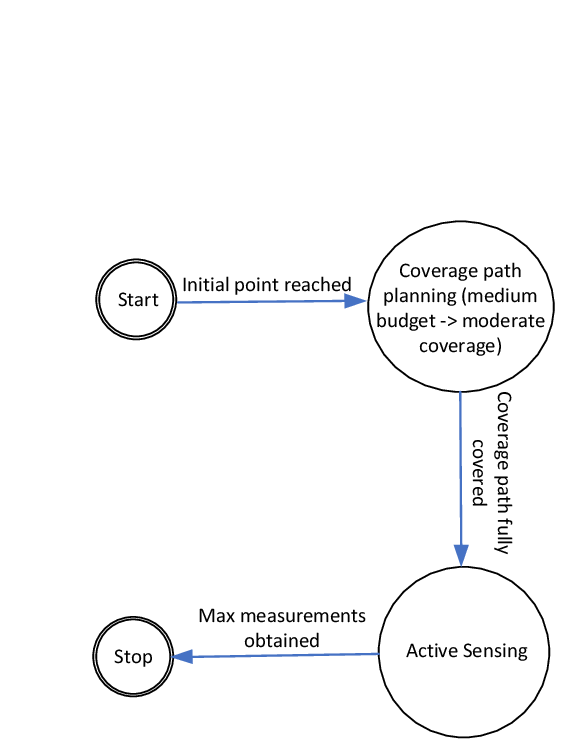}
		\caption{\textit{Specific model used for our experiments.}}
		\label{fig:reduced_state}
	\end{subfigure}
	\caption{\footnotesize  State machine diagrams: (a) general model for continuous, multi-modal switching (b) specific switching model studied in this paper.}
\end{figure}

Fig. \ref{fig:full_state} shows a general multi-model switching strategy as a state machine. This model shows the logic by which we combine budget constrained CPP \cite{tran2023coverage,perera2023radio} with active sensing. Budget constrained CPP produces a longer, more detailed coverage path when there is enough time budget, and a shorter path when there is limited time. This leads to a coarser coverage performance in the latter case, which is useful for rapid early reconnaissance. The three states on the left of Fig. \ref{fig:full_state} show proposed conditions for switching between low budget and high budget CPP, starting with the former and transitioning to the latter if nothing is found during a coarse search. We envisage that these two states may be expanded into more states if it is desired to progress more gradually to high budget CPP. 

The transitions spanning from the left to the right of Fig. \ref{fig:full_state} show the conditions under which switching from CPP to AS might be done. These include detection of a certain level of contaminant or having completed a first pass of the CPP. The transitions spanning right to left show the conditions under which switching from AS back to CPP might be done. This includes a lack of detected contaminants, which could trigger another coverage repetition or coverage with a higher budget.

Fig. \ref{fig:reduced_state} shows a simplified state machine that we study in this paper. This model makes specific choices from the general model. Specifically, this model assumes CPP with a moderate budget is completed first and then AS is triggered once the CPP has finished. We do not consider any cases that switch from AS back to CPP in this paper, but do discuss this as a direction for future work in Section V. 

The next sections now describe the budget-constrained CPP approach that we use \red{(which is adapted from \cite{tran2023coverage,perera2023radio})}, and our AS approach for cluttered environments, which is adapted from a published state-of-the-art version that was designed for static uncluttered environments \cite{tran2023coverage,perera2023radio}.

Our CPP approach is described in Section\ref{sub:MCPP} and our novel AS approach in Section \ref{sub:MRSLAS}. 

\subsection{Multi-robot coverage path planning} \label{sub:MCPP}
In many situations, there will be practical constraints on the traversal time budget in addition to wanting to maximise all information (mean and uncertainty). Adding the traversal time budget to the multi-robot coverage path planning requires further consideration, as explored in this subsection.


\subsubsection{Discretizing obstacles as a grid}
We discretize the obstacle geometry as a grid in the map. A scanline algorithm \cite{clark1979} is used to classify the grid cells as either an obstacle or a free space. 

\subsubsection{Creating Variable Sized Blocks} \label{sub:CVSB}
A block-building algorithm is applied to group adjacent free grid cells into variable-sized blocks. The largest block size is chosen to accommodate the size of the group of UGVs when they are spread out. The remaining block sizes are calculated by progressively halving the largest size. After scanning the entire map for the fitting block sizes, a list of blocks covering the entire map is obtained. A graph is built by marking the connectivity of each block with the neighbouring blocks $\{TOP, LEFT, BOTTOM, RIGHT\}$. A minimum spanning tree algorithm is applied to find a spanning tree. Connections that do not belong to the spanning tree are removed. 

\subsubsection{Minimum Spanning Tree} \label{sub:MST}
A spanning tree is a subset of an un-directed graph that connects all the vertices of the graph with a minimum number of edges. The cost of the spanning tree is the sum of edge weights in the tree. The most efficient spanning tree can be found by a minimum spanning tree (MST) algorithm. The MST algorithm constructs a tree including every vertex, where the sum of the weights of all the edges in the tree is minimized.

\subsubsection{Coverage Path Planning Algorithm}
The path planning is implemented by dividing each block into four logical parts. The center of the four parts is the connection point located on the constructed route. There are three cases for building the coverage path, depending on the number of adjacent blocks.
\begin{itemize}
   \item No adjacent block in a given direction: The centers of the two parts are simply connected.
   \item A unique adjacent block: The algorithm ignores it if the adjacent block is previously connected with this block. Otherwise, two parts of the considered block are connected to adjacent parts of the adjacent block.
   \item Two or multiple adjacent blocks: The adjacent blocks will be sorted in a direction, and then the joint points of the adjacent block pairs will be calculated by lines 16-20 of Algorithm \ref{alg:pp_algo}. The adjacent parts of the adjacent blocks will be connected to the generated joint points.
\end{itemize}
   
The above steps are illustrated in the Algorithm \ref{alg:pp_algo} in detail. Denote $D_{d,b}$ as the list of neighbor blocks of the block $b$ in the direction $d$, $Q_{c,b}$ as the coordinate of the center of the part $c \in \{TL, TR, BL, BR\}$ of the block $b$, $P_{C,b}$ as the coordinate of the center of the block $b$, $P_{L,b}$ as the coordinate of the left edge of the block $b$, and $N_b$ as the list of neighbour blocks of the block $b$.

\begin{algorithm}
\setcounter{algorithm}{1}
\caption{Multi-Robot Coverage Path Planning Algorithm \cite{tran2023coverage}} -
\label{alg:pp_algo}
\begin{algorithmic}[1]
\State \textbf{Inputs}: List of blocks $B_k$, spanning tree $ST = \{E_{B_i,B_j} | i \neq j, i,j=1,\dots,N_B\}$, starting point $p_{start}$
\State \textbf{Outputs}: List of path elements $P$.
\For{$b \in B_k$}
    \State $b_{LEFT} \gets D_{LEFT,b}$
    \If{$|b_{LEFT}|>0$}
        \If{$|b_{LEFT}|=1$}
            \If{$|D_{RIGHT,b^1_{LEFT}}|=1$}
                \State Add path $(Q_{TL, b}, Q_{TR, b^1_{LEFT}})$ to $P$
                \State Add path $(Q_{BL, b}, Q_{BR, b^1_{LEFT}})$ to $P$
            \EndIf
        \Else
            \State Sort $b_{LEFT}$ by $x$ in ascending order
            \State $n \gets |b_{LEFT}|$
            \State Add path $(Q_{TL, b}, Q_{TR, b^1_{LEFT}})$ to $P$
            \For{$i=1,\dots,n-1$}
                \State $middle \gets \frac{P_{C,b^i_{LEFT}}+P_{C,b^{i+1}_{LEFT}}}{2}$
                \State $center \gets P_{C,b}$
                \State $joint_x \gets P_{L,b}+\frac{BS_b}{4}$
                \State $joint_y \gets middle_y + \frac{(joint_x, middle_x)*(center_y-middle_y)}{center_x-middle_y}$
                \State $joint \gets (joint_x, joint_y)$
                \State Add path $(Q_{BR,b^i_{LEFT}},joint)$ to $P$
                \State Add path $(joint,Q_{TR,b^{i+1}_{LEFT}})$ to $P$
            \EndFor
            \State Add path $(Q_{BL, b}, Q_{BR, b^n_{LEFT}})$ to $P$
        \EndIf
    \Else
        \State Add path $(Q_{TL, b}, Q_{BL, b})$ to $P$
    \EndIf
    \State Do the same for the RIGHT, TOP, BOTTOM directions
    \State Sort $P$ in order of proximity to starting point $p_{start}$
\EndFor
        
\end{algorithmic}
\end{algorithm}

\begin{figure}[H]
    \centering
    \includegraphics[scale=0.35]{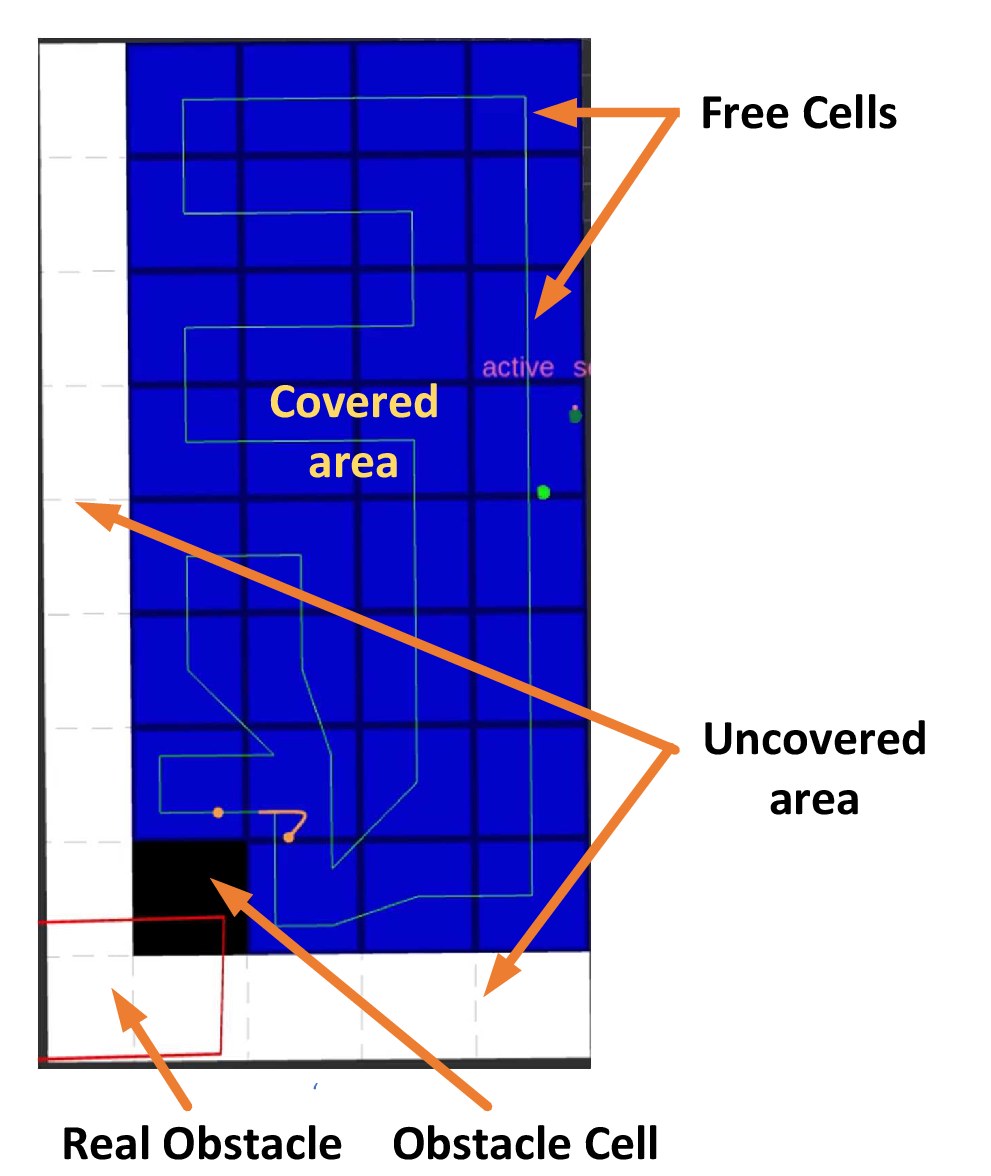}
	\caption{\footnotesize A demonstration of the CPP algorithm. The blue area is covered by the multi-robot coverage path, while the white area is missed. The black rectangular box represents an obstacle cell. Finally, the red-lined rectangular box describes a real obstacle.}
    \label{fig:cpp_area}
\end{figure}

\red{Since the dimension of the cell size represents the space covered, there are several regions where the coverage paths cannot be planned, as seen in Fig. \ref{fig:cpp_area}. The plume will be poorly resolved if these areas contain very high concentrations. Another challenge is that every location should be visited and sampled only once or twice to avoid double counting problems \cite{khaleghi2013multisensor}. As a result, the coverage paths cannot be backtracked to increase the measurement times. These issues demand an additional active sensing approach to acquiring information in the uncovered and missed areas and relaxing the detailed coverage path requirements.}


\subsubsection{Formation Control}
The block size determines the formation pattern of the UGVs ($q_1,..., and~q_5$). Virtual UGV $q_0$ serves as the leader, with the other UGVs as followers. Connecting virtual UGV $q_0$ and UGV $q_i$ ($i \geq 1$) through virtual springs (VSs) generates spring forces between them. A relative position-based formation control is derived from maintaining the desired formation shape of the UGVs \cite{tran2020switching,tran2020switching1,tran2020distributed,tran2021hybrid,tran2021multi,abpeikar2022tuning}. The spring lengths $l_0 = \{l_{01},...,l_{0n_q} \}$ are set to meet the geometrical requirement of the formations. Depending on the block size, the UGVs can take three formations: V-formation, U-formation, and Q-formation (line formation). When the block size is greater than one, V-formation or U-formation is selected. If the block size is 1, the Q-formation is chosen. Based on the block size and formation type, the desired relative positions of each follower with respect to the virtual leader are calculated to create an obstacle-avoiding formation. Each ground vehicle's control input is the resultant of the force vector produced by the virtual spring pairs. \red{This control input ensures that the followers effectively track the virtual leader when the leader robot follows the designated waypoints, thus guaranteeing the maintenance of the formation throughout the trajectory.}

\subsection{Multi-robot swarm-led active sensing} \label{sub:MRSLAS}
Algorithm \ref{alg:particle_sharing} recursively approximates the PDF to construct an unknown gas environmental model by letting multiple agents explore, given arbitrary measurement locations $l_k$, $k = 1,..., N_l$, where $N_l$ is the maximum number of measurements. In this subsection, we consider adaptively selecting the next measurement location $l_{k+1}$ based on the previous measurements $r_{1:k}$ collected so far, known as active sensing, with the purpose of obtaining faster convergence of estimating the Gaussian field parameters. The active sensing approaches of \cite{mallick2012detection,leong2021estimation,leong2022field} are extended to work for the swarm of multiple agents. 

Let $D(l^{*}, P(\xi|r_{1:k};l_{1:k})) \geq 0$ represent the reward function obtained by observing the measurement vector $r_{1:k}$ with the measurement location vector $l_{1:k}$ when the current posterior PDF is $P(.)$. Since the next measurement location $l_{k+1}$ must be chosen without the direct measurement, the objective function is to maximise a one-step ahead expected reward function (the amount of acquired information):
\begin{equation}\label{eq:point_selection}
l_{k+1} = \argmax_{i^{*} \in L_k} E[D(l^{*}, P(\xi|r_{1:k};l_{1:k}))].
\end{equation}

After the measurement is done, the quantity of information can be computed using the difference between the future posterior PDF $P(\xi|r_{1:k+1};l_{1:k+1})$ at iteration $k+1$ and the prior PDF $P(\xi|r_{1:k};l_{1:k})$ at iteration $k$. Here, the R{\'e}nyi divergence \cite{renyi1961measures} is used to measure this difference. For $\beta \in \mathbb{R} \setminus \{0,1\}$, the reward function is then given by:
\begin{equation}
E[D(l^{*}, P(\xi|r_{1:k};l_{1:k}))] \approx \frac{1}{\beta - 1}~ln~\frac{\zeta_\beta (r_{k+1}|l^{*})}{\Big(\zeta_1 (r_{k+1}|l^{*})\Big)^\beta},
\end{equation}
where 
\begin{equation}
\zeta_\beta (r_{k+1}|l^{*} \triangleq \frac{1}{N} \sum_{j=1}^{N} P(r_{k+1}|\xi_k^{(i)} ;l^{*})^\beta.
\end{equation}

The expected reward can then be derived as follows:
\begin{equation}
E[D(l^{*}, P(\xi|r_{1:k};l_{1:k}))] \approx \frac{1}{\beta - 1}~\sum_{r_{k+1} = 0}^{1} \zeta_1 (r_{k+1}|l^{*})~ln~\frac{\zeta_\beta (r_{k+1}|l^{*})}{\Big(\zeta_1 (r_{k+1}|l^{*})\Big)^\beta}.
\end{equation}

In (\ref{eq:point_selection}), a set of possible future positions $L_k$ needs to be determined. We assume that each agent can only move a limited distance from its current position $l_k$ to arrive at a new measurement location. Moreover, this set is finite for computational tractability in problem (\ref{eq:point_selection}). Denote $A$ as the allowable search area for the agent, similar to \cite{leong2021estimation}, we have:
\begin{equation}\label{eq:alex_active_sensing}
\begin{split}
L_k = \{l_k&+(j\Upsilon cos(\frac{2\pi n}{N_d}),j\Upsilon sin(\frac{2\pi n}{N_d})), \\
&\forall j = 0,..., N_{s}; n = 0,..., N_{d-1}\Big\}\bigcap A,    
\end{split}
\end{equation}
where $\Upsilon$ stands for the radial step size, while $N_{s}$ indicates the maximum number of step sizes between two successive measurement locations. Further, $N_{d}$ is the number of directions the agent can move in the plane.

\begin{figure}
	\centering
	\includegraphics[width=20.0pc]{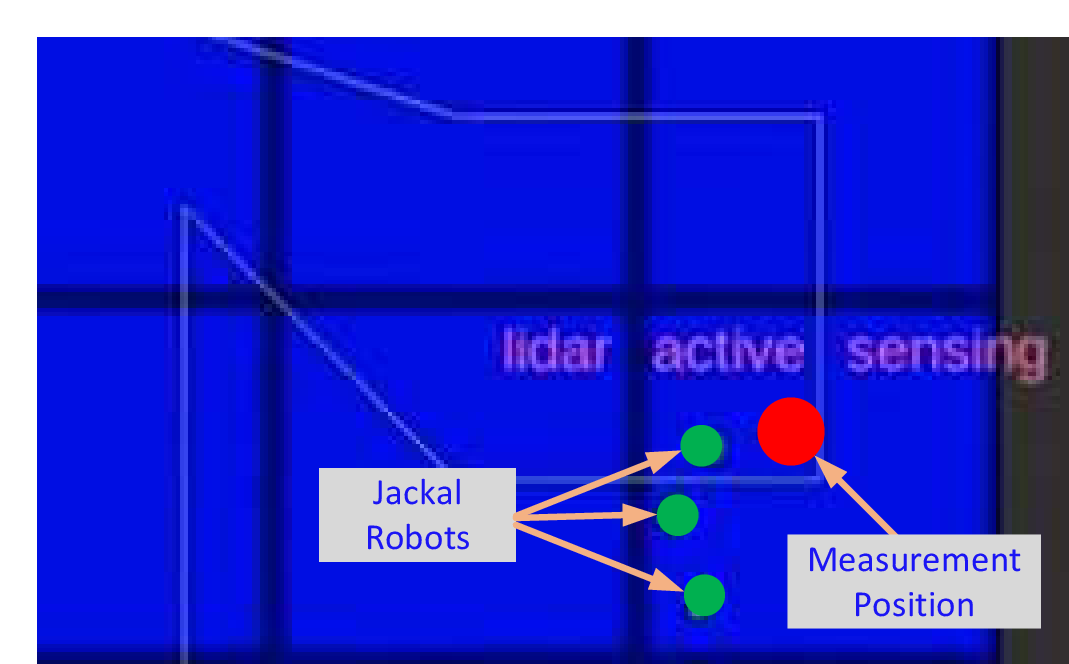}
	\caption{\footnotesize \red{Selection of the next measurement position (highlighted by a red circle) by a swarm of three Jackals (represented as green circles). Image extracted from our base station Graphical User Interface during a live trial.}}
	\label{fig:active_sensing1}
\end{figure}

Despite recent advancements in active sensing mechanisms, such as those described in \cite{leong2021estimation,leong2022field} and Equation (\ref{eq:point_selection}), gas field estimation remains a challenging task in large-scale arenas due to the varying measurement times, locations, and values resulting from each robot's independent sensing actions. In this work, we propose a novel solution to address these challenges, whereby each robot in the swarm broadcasts its highest reward value and corresponding location to enable all robots to select the same final measurement point which has the highest reward value (see (\ref{eq:fused_point})) and Fig. \ref{fig:active_sensing1}. By doing so, our approach ensures more consistent and accurate gas field estimations, offering a significant improvement over existing methods:
\begin{equation}\label{eq:fused_point}
l_{k+1} = \argmax_{k=1}^{N_r} E(.)^{(k)},
\end{equation}
where $N_r$ is the number of unmanned ground vehicles.

To ensure the safety of the robot swarm and prevent collisions with obstacles, it is important to address two key issues. First, the robot flock must be able to avoid any mobile obstacles that might obstruct its path to the destination. This can be achieved by implementing the closest safe angle-based obstacle avoidance strategy, as discussed in Section \ref{sub:MRSLAS}. Second, to avoid the robot swarm selecting the next measurement point within an obstacle, Equation (\ref{eq:alex_active_sensing}) must be adjusted to consider only the free-obstacle area $F$ surrounding the robot swarm while excluding the area of obstacles $OB$ located in $A$. The modified equation is shown below:
\begin{equation}\label{eq:new_active_sensing}
\begin{split}
L_k = \{l_k&+(j\Upsilon cos(\frac{2\pi n}{N_d}),j\Upsilon sin(\frac{2\pi n}{N_d})), \\
&\forall k = 0,..., N_{s}; n = 0,..., N_{d-1}\Big\}\bigcap F, \\
F = A &- OB,~\forall F,~OB \in A.  
\end{split}
\end{equation}

Here, $F$ represents the free space within $A$, while $OB$ denotes the obstacle area within $A$. In addition, an exploration parameter $\theta$ is introduced such that with probability $\theta$, the next waypoint will be randomly selected within the search space $F$. Conversely, with probability $1 - \theta$, the optimization in Equation (\ref{eq:fused_point}) is carried out. This strategy is reminiscent of the exploration-exploitation trade-off in reinforcement learning \cite{ji2021multi,leong2021estimation}.

The AS algorithm is depicted in Algorithm \ref{alg:c_active_Sensing}.

\begin{algorithm}
\caption{AS: $x'_{t+1}$ = Collaborative Entropy-Driven Active Sensing($x_t^{(r)}$,$\xi^{j}$)}
\label{alg:c_active_Sensing}
\begin{algorithmic}[1]
\State \textbf{Algorithm Parameters}: $\beta \in [0,\infty)/\{1\}$, $\varepsilon \geq$ 0, $\Upsilon \geq$ 0, $N_{s} \in \mathbb{N}$, $N_d \in \mathbb{N}$, search region $A$, free space in search region $FS$, obstacle region in search region $OB$, $\mathcal{N}_r \in \mathbb{N}$
\State \textbf{Inputs}: $x_k$, $\xi_t^{(j)}$
\State \textbf{Outputs}: Next measurement location for the flocking multi-robot system $x'_{t+1}$
\State With  probability $\varepsilon$ of exploration, and $1 - \varepsilon$ of exploitation, if the algorithm falls in the exploration phase, set $x'_{t+1}$ to a random location in $F$ and $E$ to $\infty$, otherwise set:\[\
l_{k+1} = \argmax_{i^{*} \in L_k} \frac{1}{\beta - 1}~\sum_{r_{k+1} = 0}^{1} \zeta_1 (r_{k+1}|l^{*})~ln~\frac{\zeta_\beta (r_{k+1}|l^{*})}{\Big(\zeta_1 (r_{k+1}|l^{*})\Big)^\beta}
\]
where \[
L_k = \{l_k + (j\Upsilon cos(\frac{2\pi n}{N_d}),j\Upsilon sin(\frac{2\pi n}{N_d})), \]
\[
\forall k = 0,..., N_{s}; n = 0,..., N_{d-1}\Big\}\bigcap F, \]
\[
F = A - OB,~\forall F, OB \in A. 
\]
\State Broadcast $x_{t+1}^{(k)}$, $\mathcal{E}^{(k)}$ to all robots flocking.
\State Determine the final destination for the multi-robot system:
\[
l_{k+1} = \argmax_{k=1}^{N_r} E(.)^{(k)}
\]
\end{algorithmic}
\end{algorithm} 

Although the proposed active sensing method can maintain connectivity of the communication network between robots and the robots are able to construct the gas field with reasonable accuracy, the time it takes to converge is often inconveniently long, which hinders its use in real-time tasks where the completion time is critical, or robot batteries only last for a limited time. The main reasons are the shortage of information and the high mismatches between our model assumptions and realistic gas dispersion in initial locations, resulting in the incorrect measurement positions of the robots.

\begin{figure}
	\centering
	\includegraphics[width=24.0pc]{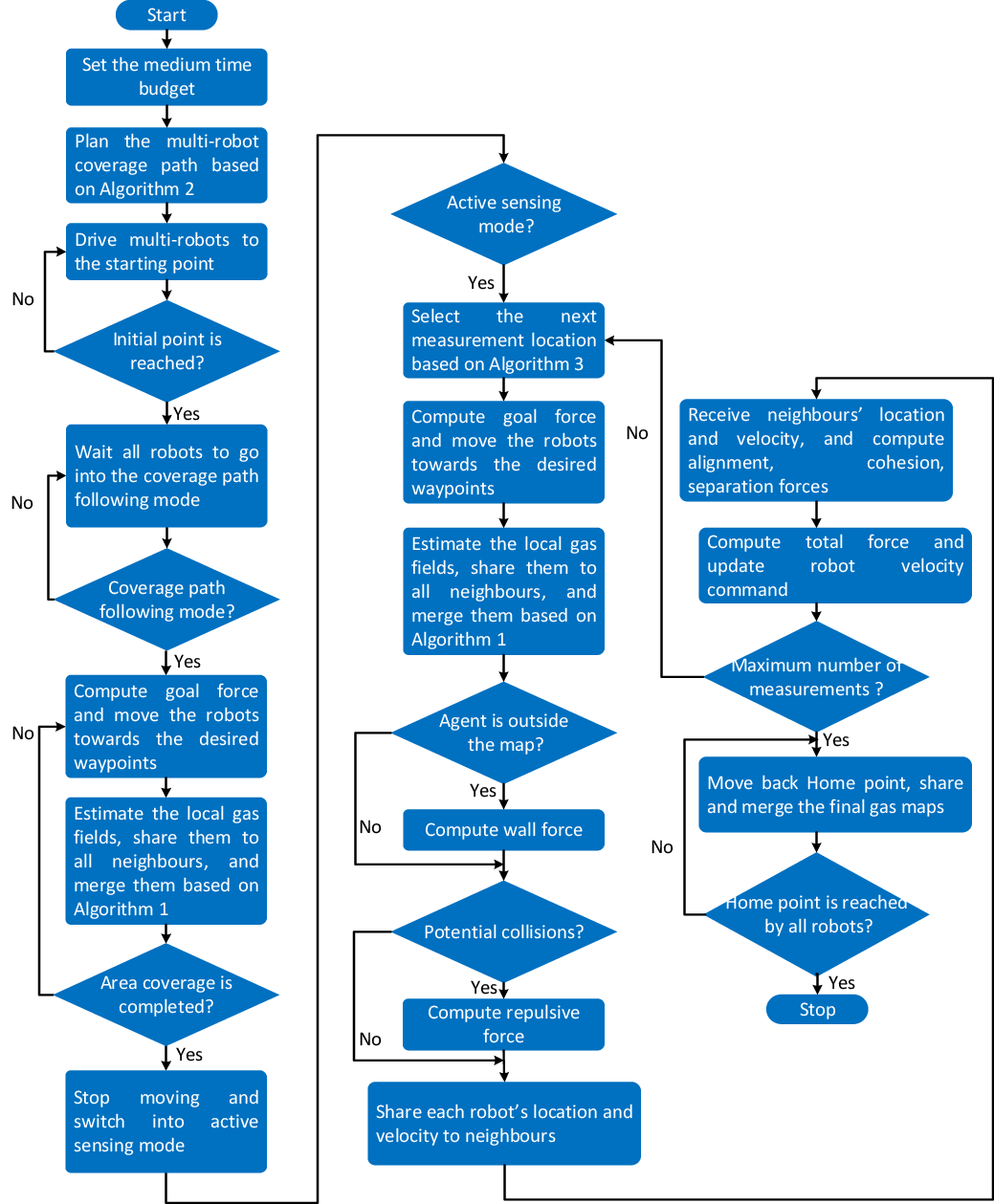}
	\caption{\footnotesize \red{Flow chart of the proposed switching exploration algorithm.}}
	\label{fig:flowchart}
\end{figure}

The overall procedure of the whole proposed control algorithm is depicted in Fig. \ref{fig:flowchart}.

\section{Real-Time Experiments}
In this section, we begin with a description of our hardware-in-the-loop and real-time experiments in Part A. We then describe the metrics we use to measure performance in Part B and summarize the results of our experimental studies in Part C.

\subsection{Experimental setup}

\subsubsection{The environment}
\begin{figure}
	\centering
	\includegraphics[width=23.0pc]{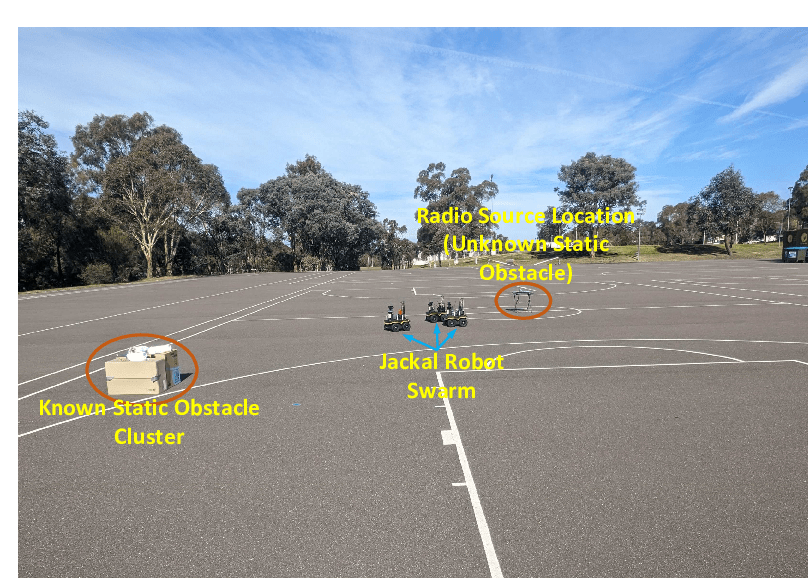}
	\caption{\footnotesize Experimental setup. The radio source is placed close to the middle of the mapped area, and its location is measured using a GPS sensor.}
	\label{fig:jackals}
\end{figure}

The experimental setup was designed to replicate a real-world gas source localization and mapping scenario in an obstacle-rich arena, measuring 14.98m $\times$ 28.12m. This arena featured two distinct categories of obstacles with arbitrary geometry. The first category involved a set of known complex C-shaped static obstacles (consisting of four obstacles) deliberately positioned in the lower-left corner. This area, known for its highest gas concentration, was frequently traversed by all robots during the active sensing phase. The second category introduced an unpredictable element by incorporating an unknown obstacle (a table in our experiment), deliberately located at the gas source position  (see Fig. \ref{fig:jackals}). This deliberate obstruction in the coverage and active sensing paths demanded obstacle avoidance in the exploration phase and no measurement point selection within the obstacle area in the active sensing phase, hindering gas measurements at the source location. Such scenarios, mirroring real-world complexities, were deliberately selected to enhance the experimental fidelity. Finally, the environment was made more challenging with the inclusion of dynamic obstacles, (a human in our experiment) moving in an unpredictable manner. This dynamic obstacle had the potential to obstruct the robots' planned paths, introducing an additional layer of realism to the experimentation.

\begin{figure}
	\centering
	\includegraphics[width=14.0pc]{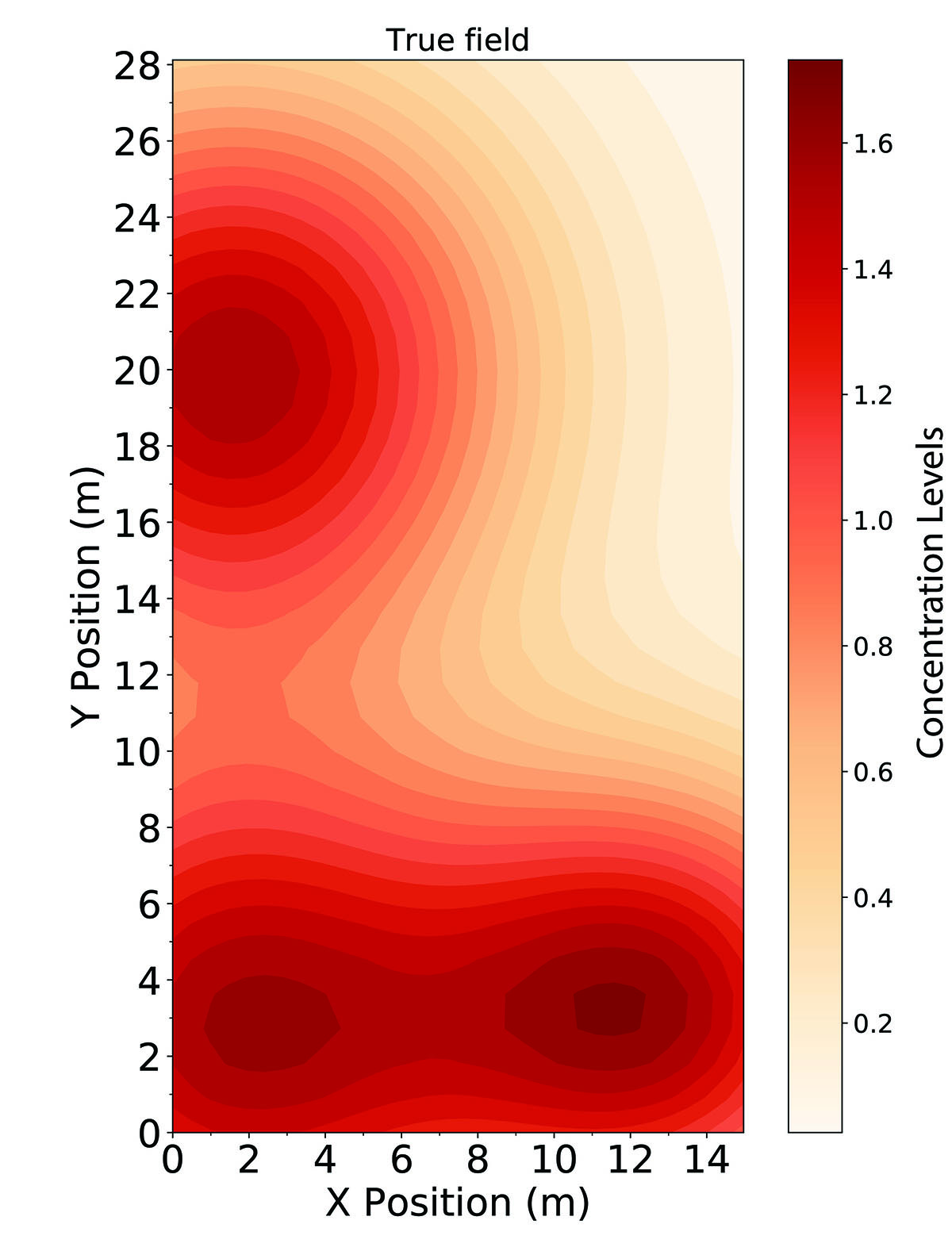}
	\caption{\footnotesize True gas field concentration contours.}
	\label{fig:true_field}
\end{figure}

\red{Real-world evaluation of gas source localization and mapping strategies is challenging due to the unavailability of ground truth data for gas concentration distributions. To address this, we designed an initial set of hardware-in-the-loop experiments using a synthetic gas field with a physical team of three Jackal robots. While the gas field is simulated, this approach provides accurate ground truth data on source distribution and gas concentration. This setup facilitates comprehensive comparisons and demonstrates the general feasibility of the CPP-AS approach.} The scalar field is governed by the superposition of 16 Gaussian fields ($I = 16$) and is simulated for the 2D case according to Eq. \ref{eq:Gauss_field}. The true field's other parameters are as follows: $\tilde{\gamma}_1 = 1.6$, $\tilde{\gamma}_2 = 1.4$, $\tilde{\gamma}_3 = 1.6$, $\tilde{\mu}_1 = (1.6, 20)$, $\tilde{\mu}_2 = (12.8, 3.3)$,  $\tilde{\mu}_3 = (1.6, 2.7)$, $\tilde{\sigma}_1^2 = 7.7$, $\tilde{\sigma}_2^2 = 6$, $\tilde{\sigma}_3^2 = 7.7$. This way, we have ground truth data of the simulation's source distribution and gas concentration. Fig. \ref{fig:true_field} demonstrates the true field with the proposed parameters. Unlike the experiments performed in \cite{wiedemann2017bayesian,leong2022field}, where only one source or robot is involved, we established three sources in the environment and strategically positioned the gas sources to explore various aspects of the gas distribution mapping problem. One source was placed at the back of the region and more isolated from the other two. The other two sources were placed closer to the start point of the robots, and closer to each other. The sources were strategically placed near the left and bottom boundaries, which are often partially covered by the coverage path planning strategies. This placement aimed to assess the robots' ability to locate the gas sources and map the gas field effectively in challenging scenarios near the boundaries. The turnaround time the multi-robot system takes to locate multi-leak sources successfully is significantly longer than the time to locate a single leak, which presents another challenge.


For field estimation, the measurement noise variance is $\tilde{\sigma}_v^2 = 0.32$, and the sensor threshold is $\tau = 1$.  We consider the case where $\mu_i$, and $\sigma_i^2$ are set as in Section \ref{sub:GPM}, while $\alpha_{i}$ and $\sigma_v^2$ are identified by the collaborative particle fusion approach.\\

\red{For our second set of experiments, we replaced the simulated gas plume with a radio source detected by a radio sensor on each robot to evaluate our approach in more realistic settings}. Specifically, we used low-power S6B model XBee Wi-Fi modems as transmitters and receivers, which support RF data rates up to 27 Mbps. These are equipped with whip antennas that protrude about 25 mm above the PCB surface, as demonstrated at the bottom of Fig. \ref{fig:hardware}.


To conduct the real-time experiments, we placed one XBee modem near the middle of our environment to serve as the signal transmitter (located at (6.73m,12.80m)). The other modems were mounted on three Jackal robots (see Fig. \ref{fig:jackals}). During the experiment, each XBee modem measures the Received Signal Strength Indicator (RSSI) value and transmits the RSSI readings to each relevant robot via Robot Operating System (ROS) messages at a fixed frequency of 1 Hz. The RSSI values remained within a range of 15dBm to 50dBm, removing any values outside of this range. Additionally, we applied a moving average filter with a window size of 5 samples to smooth out any short-term overshoots or noisy fluctuations in the RSSI values.

\subsubsection{The Robots}

\begin{figure}
\centering
\includegraphics[width=0.4\textwidth]{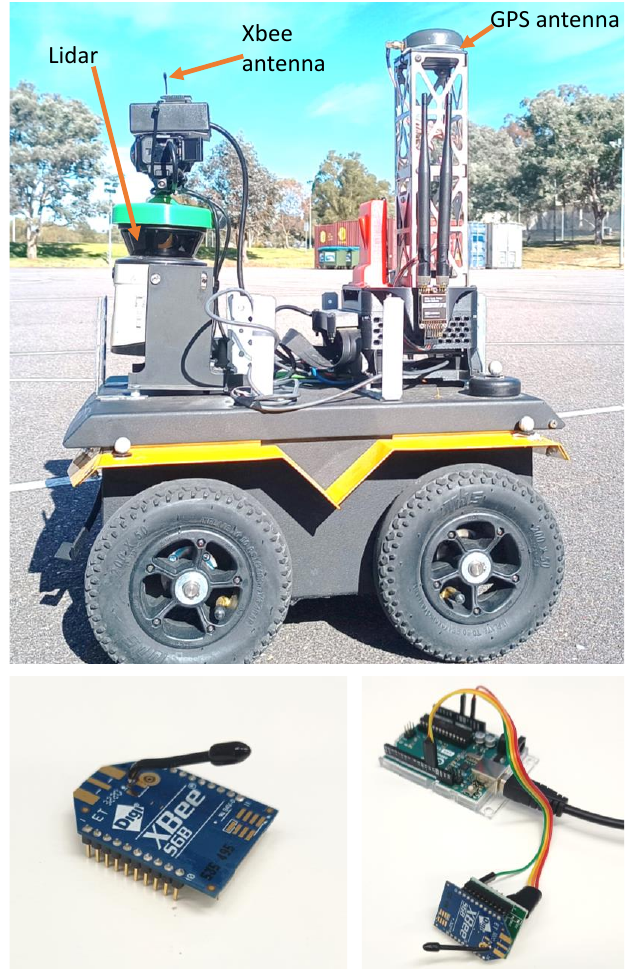}
\caption{The Jackal mobile robot used for the radio source localisation and field tests is shown on top. The XBee modems are mounted on the robot with the help of a plastic bracket. The XBees are placed as high as possible on the robot to minimize ground reflections. The XBee modem and the Arduino attachment used for the testing are shown at the bottom.}
\label{fig:hardware}
\end{figure}

We used a real team of three Jackal mobile robots, one of which is shown in Fig. \ref{fig:hardware}. Each Jackal is equipped with a differential GPS (DGPS) module, Inertial Measurement Unit (IMU) and a SICK LMS-111 LiDAR. The LiDAR’s observation range is from 0.5m to 20m. The incoming GPS signal and the internal IMU sensor are read at a rate of 10Hz. The DGPS base station was stationary and transmitted DGPS corrections to the UGVs. The sampling time of the whole system was 30Hz, except that the DGPS, IMU, and LiDAR sensors are sampled at a frequency of 10Hz. The role of the ROS Master is to enable individual ROS nodes to locate one another. Unlike traditional implementations where the ROS Master is run only on a single base station, the ROS Master was implemented on each UGV and our base station to facilitate ROS communications and improve the system's robustness. The virtual leader’s forward speed $\upsilon_l$ was initialised to 0.16m/s while the robot's maximum linear speed $\upsilon_r$ was 0.15m/s. The maximum turn rate was set to 0.73rad/s. The starting point on the coverage map is located at (10.559m,21.77m). Other parameters of our numerical experiments are summarized in Table \ref{tab:est_param}.

We found that the Gaussian model-based estimation should be limited to an update rate of 0.2 Hz (5s for one update). At higher update rates, the communication network can become congested. Further, the Jackal battery lasts only 2 hours, limiting us to approximately 70 measurements for each trial.

\begin{table}[htbp]
 \centering
\caption {Algorithm parameters used in the first experiment and their values adapted from \cite{tran2023multi,tran2023dynamic}.} 
\label{tab:est_param}
  \centering
  \begin{tabular}{lll}
    \hline
     \textbf{Parameter} &  \textbf{Description} &  \textbf{Value} \\
      \hline
$I$ &  Number of basis functions & 16 \\
$N_l$ &  Number of measurements  & 359 \\
$N_p$ &  Number of particles & 5000 \\
$N_{s}$ &  Maximum Number of step lengths & 1 \\
$N_d$ & Number of the heading robot directions & 10  \\
$\epsilon$ &  Exploration probability & 0.01  \\
$\omega_{c}$ &  Weight for cohesion rule & 0.6 \\
$\omega_{a}$ &  Weight for alignment rule  & 0.5 \\
$\omega_{s}$ &  Weight for separation rule & 1.12 \\
$\omega_{w}$ &  Wall Weight & 1.0  \\
$\omega_{at}$ &  Attractive force Weight & 2.59 \\
$R_{a}$ (m) &  Alignment Radius & 3  \\
$R_{s}$ (m)&  Separation Radius & 0.95  \\
$R_{c}$ (m)&  Cohesion and Communication Radius & 3 \\
$R_{av}$ (m)&  Obstacle Avoidance Radius & 0.5  \\
\hline
  \end{tabular}
\end{table}

All parameters for the real-source experiment were kept identical to those used in the hardware-in-the-loop exploration experiments, except for the threshold $\tau$, which was set to 35dBm for both Jackals. This threshold corresponds to the highest gas concentration level within a predefined radius of 1m, indicating the region where the gas concentration exceeds the specified threshold.

\subsection{Metrics}\label{sub:M}

In the first set of experiments, we will compare the proposed switching exploration strategy (CPP-AS) against exploration with only active sensing (AS) using the same parameters. Three evaluation metrics for the real-time experiment are presented; they are: (1) average mean squared error measured on the estimated field (ANMSE) (\ref{eq:AMSE}), (2) total exploration time, and \red{(3) averaged source position error (ASPE)}. Additionally, each experiment is repeated five times. The mean and 95\% confidence interval are computed and shown in all plots.

\begin{equation}\label{eq:AMSE}
ANMSE = \frac{1}{N_l~N_r} \sum_{k=1}^{N_r} \sum_{j=1}^{N_l} MSE_j^{(k)},
\end{equation}
where,
\begin{equation}\label{eq:MSE}
\begin{split}
MSE_j^{(k)} &= \frac{1}{\lvert A\rvert} \sum_{l \in A} \Big(\sum_{i=1}^{I} \gamma_{i} \exp(-\frac{\left\|\ l -\mu_{i} \right\|^2}{\sigma_{i}^2}) \\
&- \sum_{i=1}^{I} \hat{\gamma}_{i,t}^{(k)} \exp(-\frac{\left\|\ l -\hat{\mu}_{i,t}^{(k)} \right\|^2}{{(\hat{\sigma}_{i,t}^{(k)}})^2})\Big)^2
\end{split}.
\end{equation}

To evaluate the gas mapping results in the hardware-in-the-loop experiments, we carried out an additional real-time test in the second experiment, using the CPP-AS strategy with a single radio source. In this test, we assessed the accuracy of the estimated radio source location by calculating the distance error between the actual radio source position $l_s$ and the estimated position $\hat{l_s}$. We opted for this evaluation metric instead of using MSE as in the previous experiment since the true field's parameters are unknown:

\begin{equation}
E_d = \norm{l_s - \hat{l_s}}_{2},
\end{equation}
where $\hat{l_s}$ will be calculated by finding the position with the maximum estimated RSSI value:

\begin{equation}
\hat{l_s} = \argmax_{\forall l \in \mathbb{N}^2} (\hat{c(l)}).
\end{equation}

\subsection{Results and Discussion}

This section discusses results for: (1) the coverage path planning stage of the algorithm with the virtual gas field, (2) the path following and active sensing stage with the virtual gas field, (3) the path following and active sensing stage with a single real source, and (4) the effect of robot quantity and environment size on the performance of the CP-AS algorithm.

\subsubsection{Coverage Path Planning Phase with the Virtual Gas Field}

\begin{figure}
	\centering
	\includegraphics[width=16.0pc]{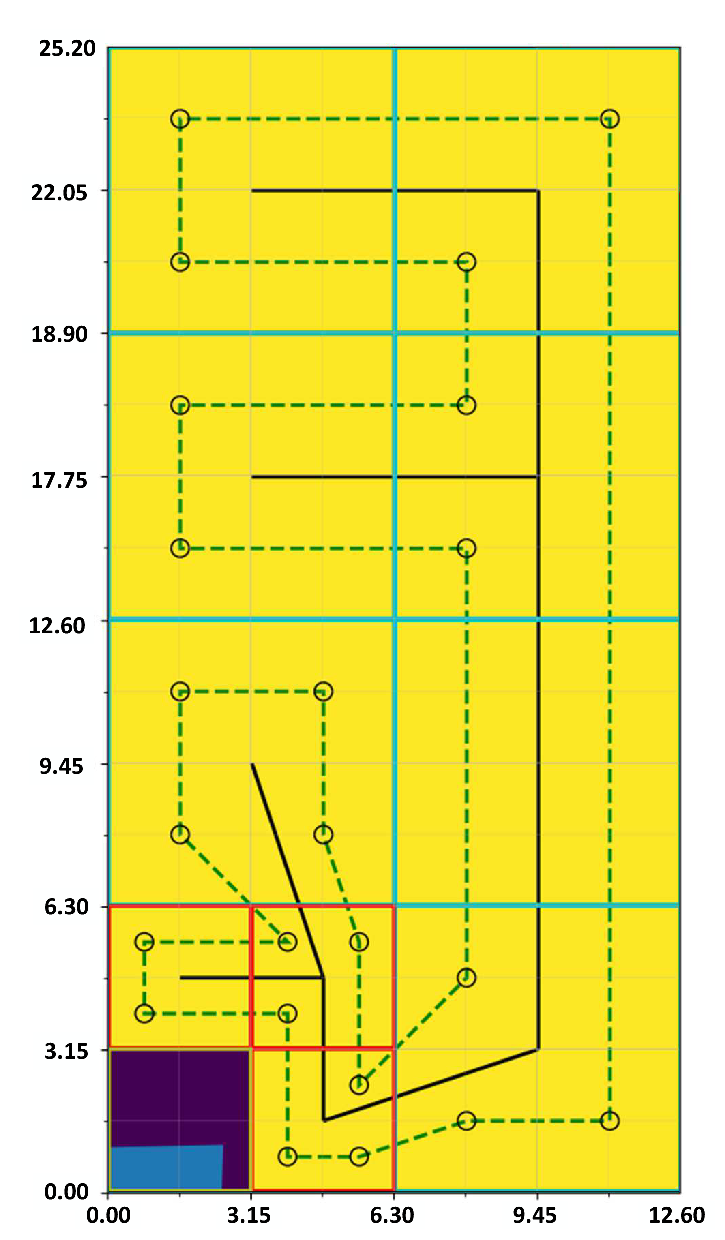}
	\caption{\footnotesize Predicted path for the virtual leader in the outdoor field site. The dark blue cells show (filled) obstacles, while the yellow cells are free space. The dashed green line represents the coverage path. The minimum spanning tree is shown in the black line. Finally, the open space is decomposed into variable-sized square blocks that follow the grid lines. These blocks can be red (2 cells for queuing formation) and aqua (4 cells for V formation)}.
	\label{fig:cpp_map}
\end{figure}

For coverage path planning, the maximum traversal time budget is set to moderate ($\bar{TT} = 1500s$ - barely enough time for the CPP algorithm to complete). Fig. \ref{fig:cpp_map} presents the robot trajectories, where the prediction engine took 2s to generate a solution. It also recommends a grid cell size of 3.15m and forecasts a complete coverage time $\hat{TT}$ of 910s and a coverage percentage of 100\%. 

\begin{figure}
	\centering
	\includegraphics[width=22.5pc]{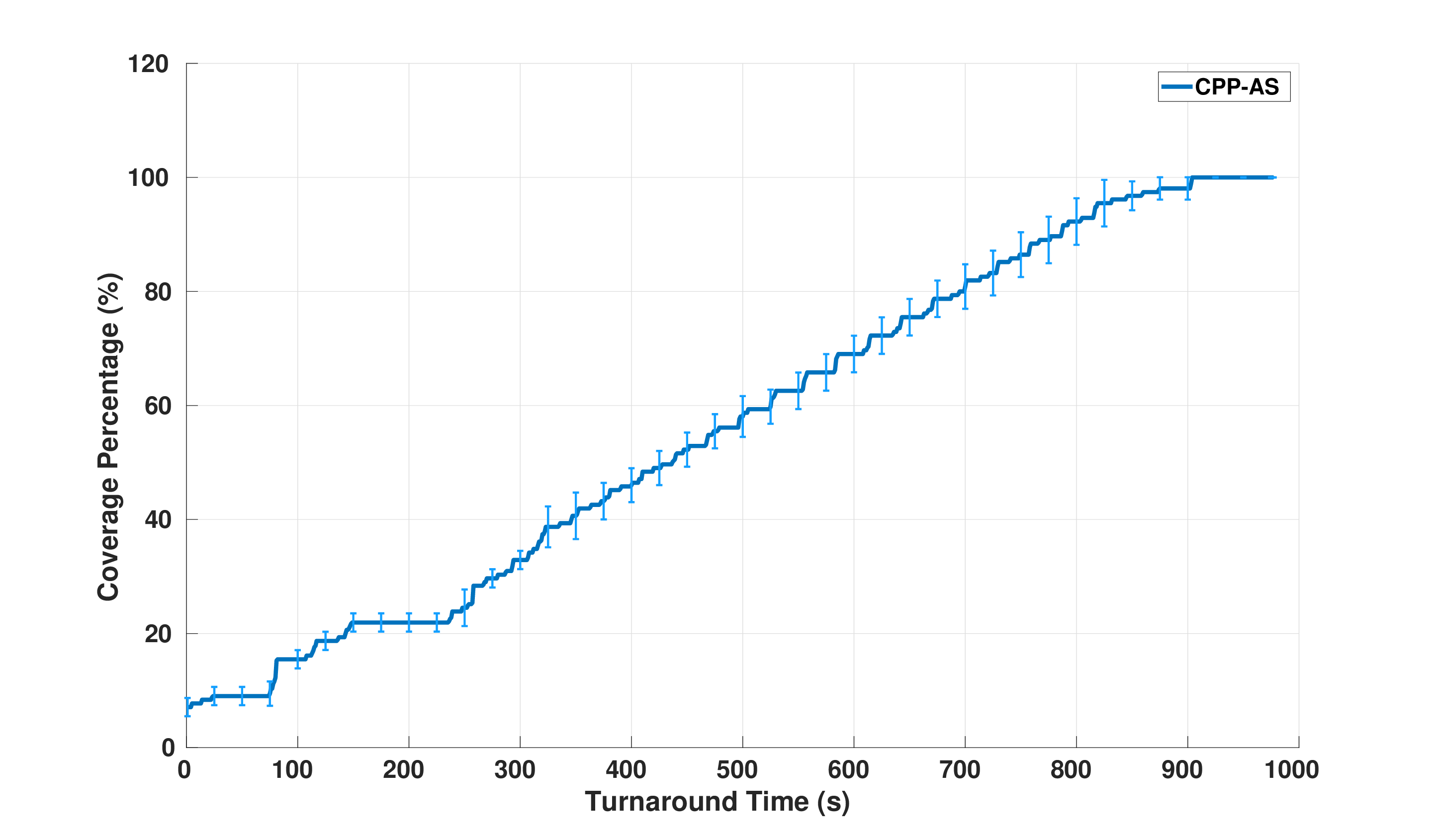}
	\caption{\footnotesize Coverage percentage over time produced by the multi-robot formation during the CPP phase.}.
	\label{fig:cp_metrics_exp2}
\end{figure}

\red{This figure reveals that the coverage space's dimension is only 12.6m $\times$ 25.2m, which is relatively smaller than the arena's full dimension that was surveyed. Consequently, there is a lack of information in several areas at the end of the CPP step. Nonetheless, as shown in Fig. \ref{fig:cp_metrics_exp2}, the robots using CPP were able to visit 96.77$\pm0.00\%$ directly and 3.23$\pm0.00\%$ indirectly after 910s. This is almost identical to the coverage time and coverage percentage estimated by the prediction engine.}

However, \red{the CPP-only approach yields an ANMSE value of 0.112$\pm$0.047, indicating a moderate difference between the true field and the reconstructed field. This disparity results from unexplored boundary areas, where three sources are located, as discussed in the concluding paragraph of Section 2.3.4. This highlights the need for an additional active sensing method to thoroughly explore and address these regions.}

\subsubsection{Active sensing results with the Virtual Gas Field}

\begin{figure}
	\begin{center}
		\begin{tabular}{cc}	
			\includegraphics[width=27.5pc]{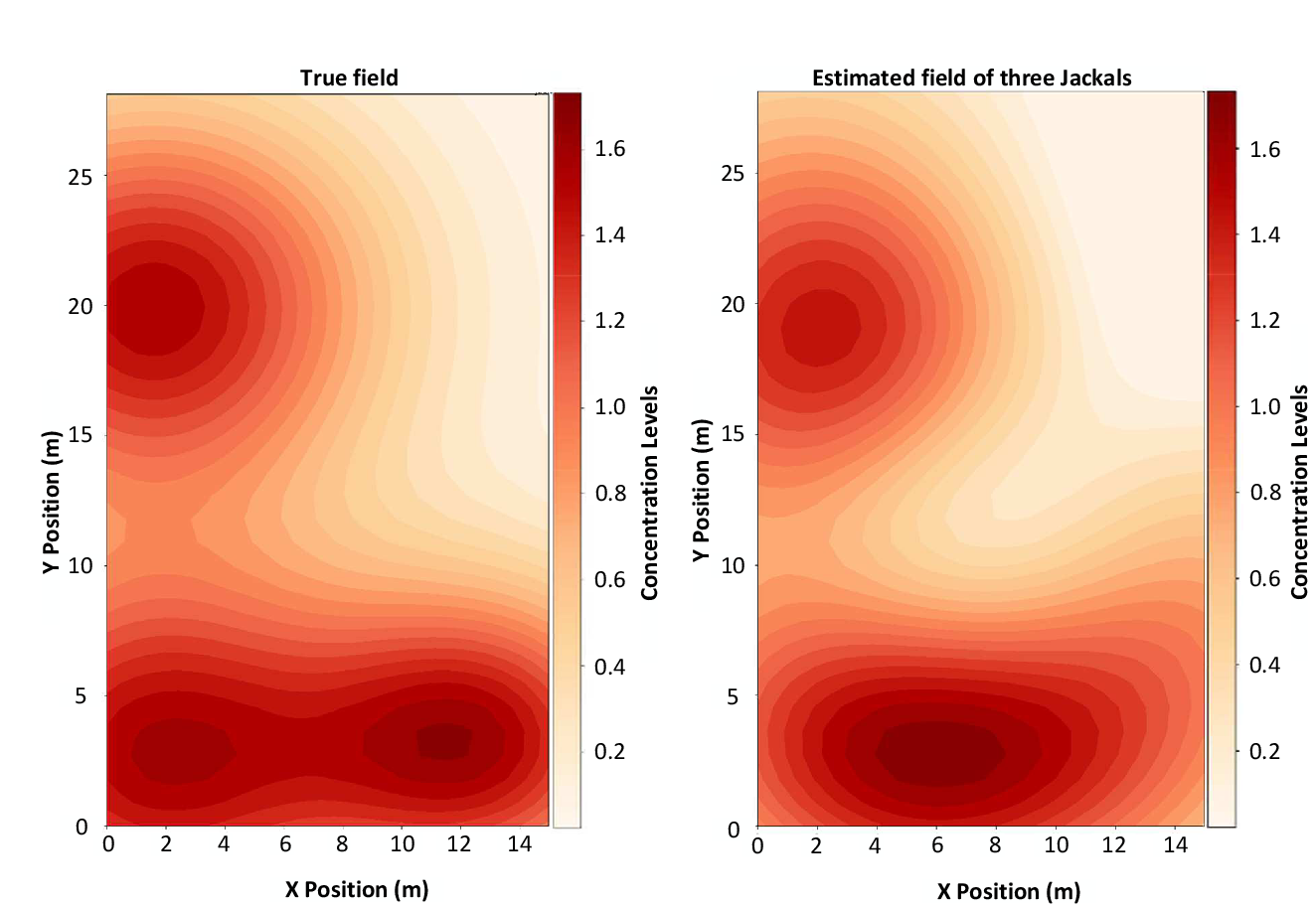} \\
			(a) \textit{CPP-AS.}\\[6pt]
			\includegraphics[width=27.5pc]{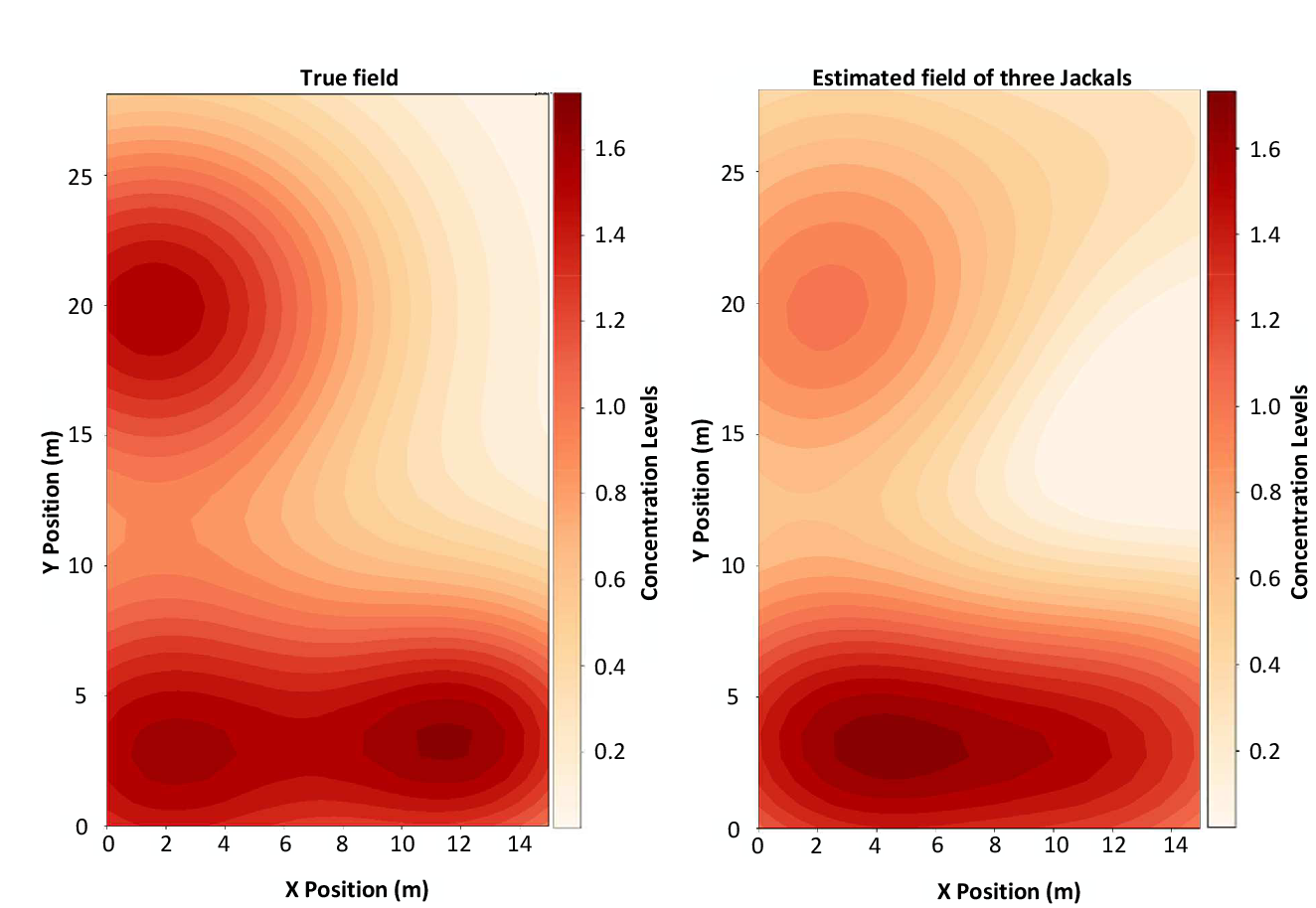} \\
			(b) \textit{AS.}\\[6pt]
		\end{tabular}
		\caption{\footnotesize \red{Sample gas maps produced by the 3 Jackals, compared to the true gas field, after 359 measurements of (a) CPP-AS and 68 measurements of (b) AS \cite{leong2021estimation,leong2022field,tran2023multi}.}}
		\label{fig:gas_maps}
	\end{center}
\end{figure}

Fig. \ref{fig:gas_maps} shows the estimated gas fields learned by each of the 3 Jackal robots when maximum measurements are obtained. We can see that the local concentration fields estimated by different robots using the CPP-AS are visually similar to the true field. There is greater visual similarity for CPP-AS than there is for AS, which does not well identify the third source towards the back of the test arena.

\begin{figure}
	\centering
        \begin{subfigure}[b]{0.495\textwidth}
		\includegraphics[width=\textwidth]{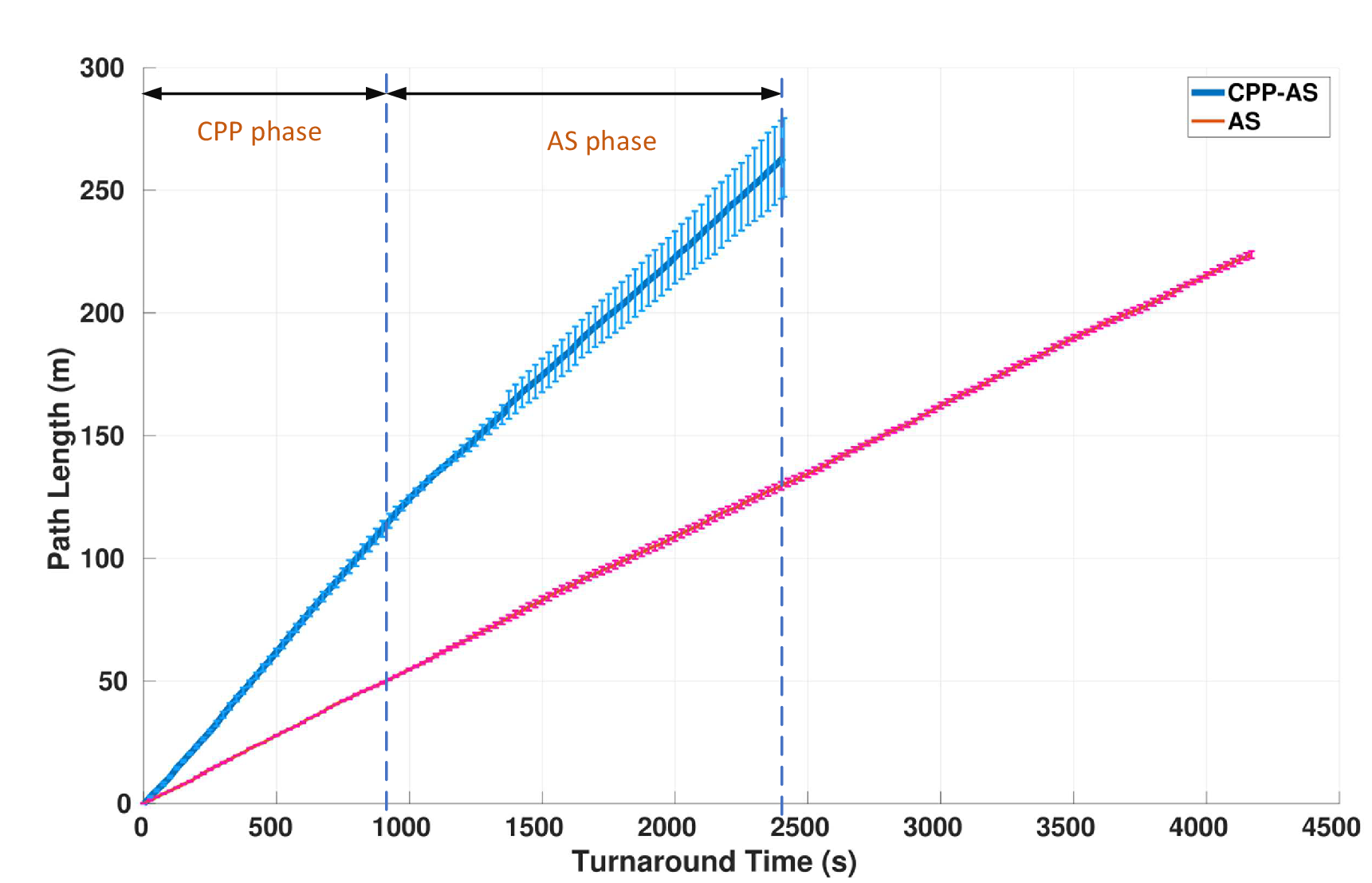}
		\caption{\textit{{Path Length vs elapsed time}}}
		\label{fig:pl2}
	\end{subfigure}
      \hfill
        \begin{subfigure}[b]{0.495\textwidth}
		\includegraphics[width=\textwidth]{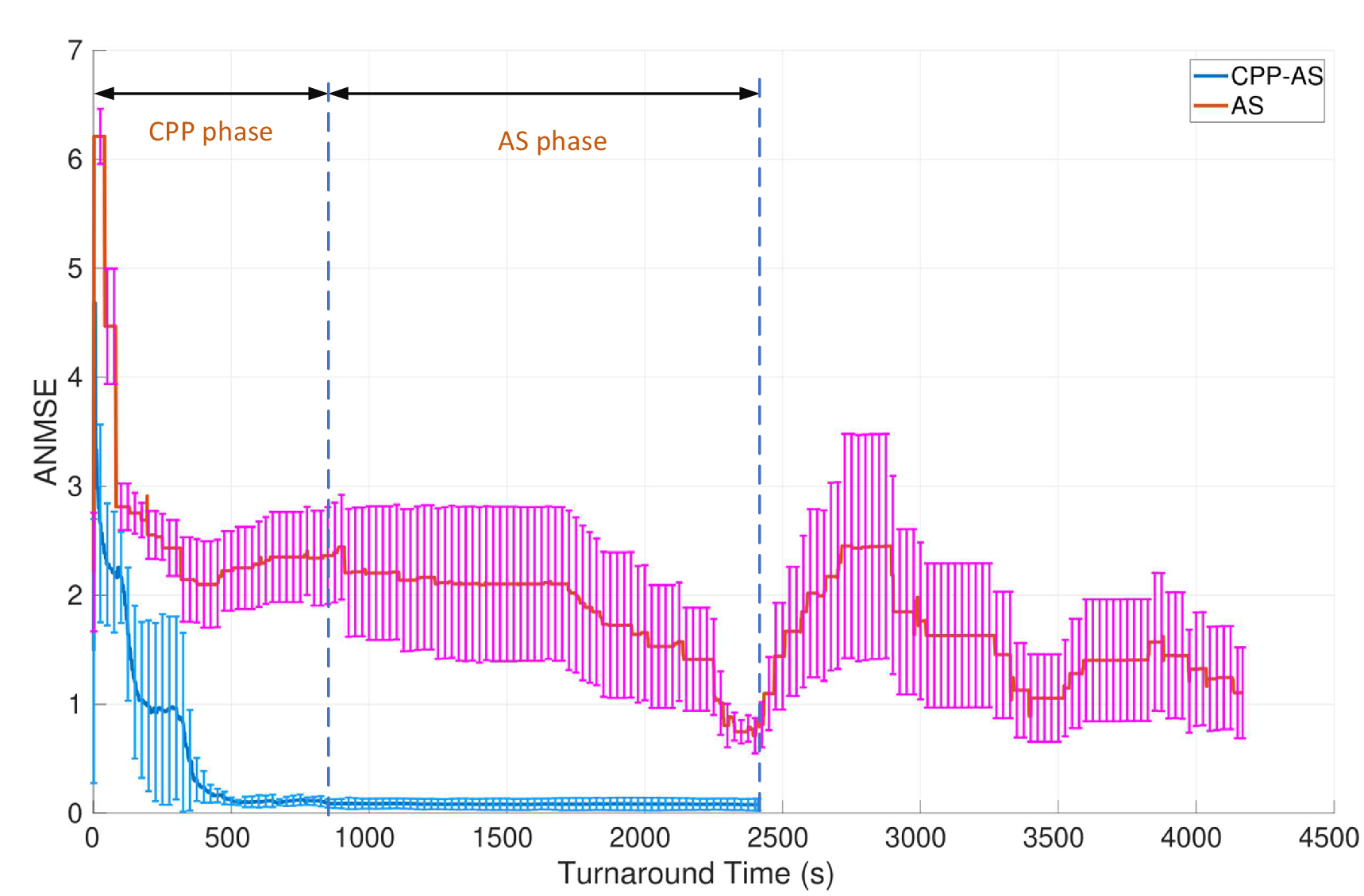}
		\caption{\red{\textit{{ANMSE vs elapsed time}}}}
		\label{fig:ams2}
	\end{subfigure} 
 	\caption{\footnotesize Hardware-in-the-loop results: The two plots compare the performance of the proposed CPP-AS and the AS strategy using the metrics from Section \ref{sub:M}.}
	\label{fig:metrics_exp1}\vspace*{-1pt}
\end{figure}

The performance of the CPP-AS and AS algorithms is evaluated in terms of their ability to estimate the gas field accurately, as measured by the ANMSE, and their efficiency in terms of traversed path length and exploration time. The results are presented in Fig. \ref{fig:metrics_exp1}. Fig. \ref{fig:metrics_exp1}(b) reveals that the CPP-AS achieves significantly lower ANMSE than the AS at the 95\% confidence level, even when both have taken 68 measurements. Although the CPP-AS covers the entire map before active sensing, leading to a relatively lengthy traversed path, (Fig. \ref{fig:metrics_exp1}(a)), it takes only 2441.54$\pm$553.64s to reach 359 measurements. 
In contrast, AS takes approximately 4328.48$\pm$437.04s to only explore 68 positions. \red{After exploring for this duration, the CPP-AS achieves a significantly lower ANMSE value of 0.062 $\pm$ 0.035 compared to that of the AS and CP-only}. Fig. \ref{fig:metrics_exp1}(a) further shows that the CPP-AS continues to yield lower ANMSE values than the AS over time and converges after just 500s compared to 3500s for AS. \red{These results demonstrate the effectiveness of the CPP-AS over the AS and CP-only in achieving more accurate and efficient gas field mapping. All results are significant at the 95\% confidence level.}

\begin{figure}
	\begin{center}
		\begin{tabular}{cc}	
			\includegraphics[width=26.5pc]{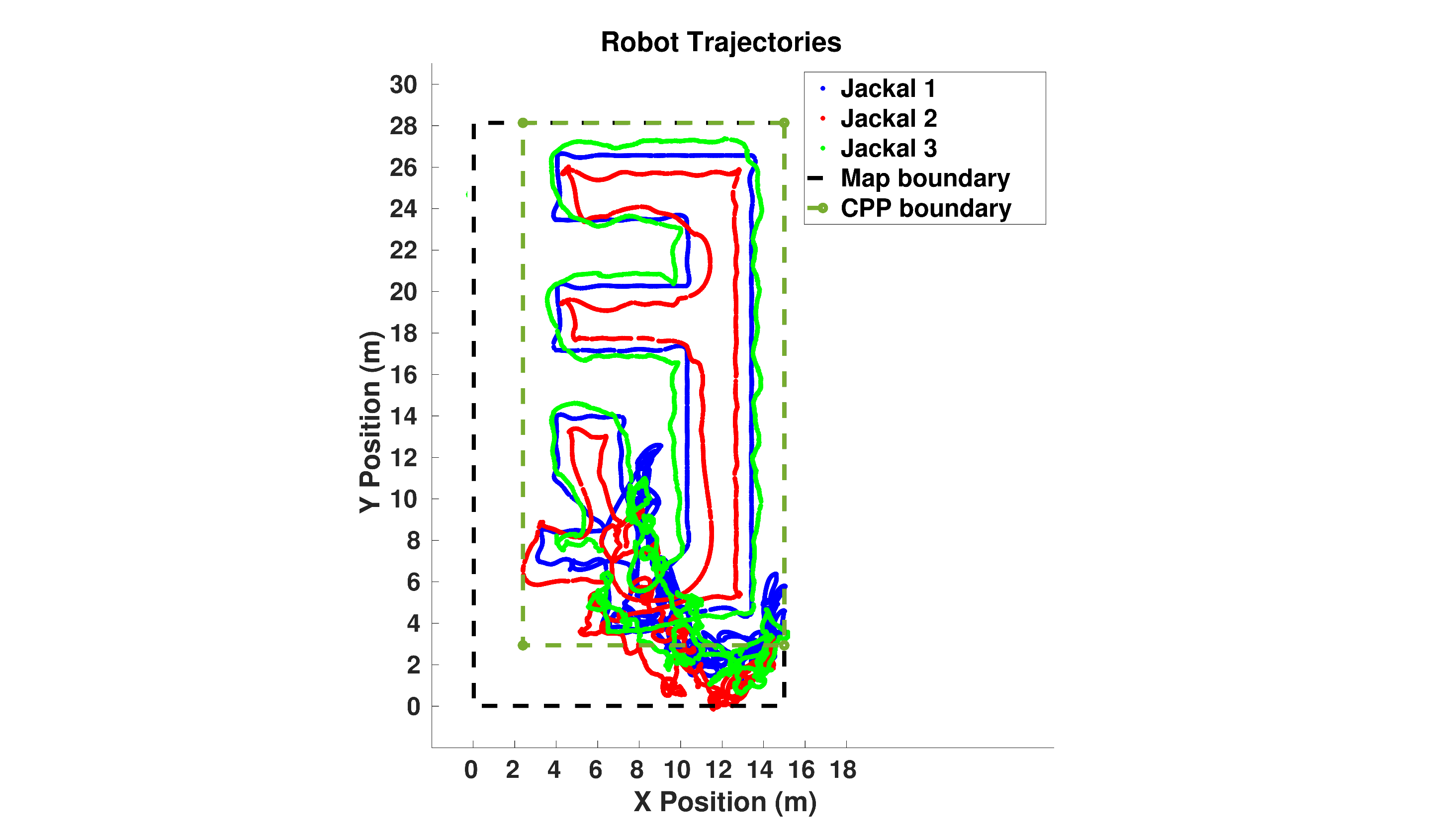} \\
			(a) \textit{CPP-AS.}\\[6pt]
			\includegraphics[width=26.5pc]{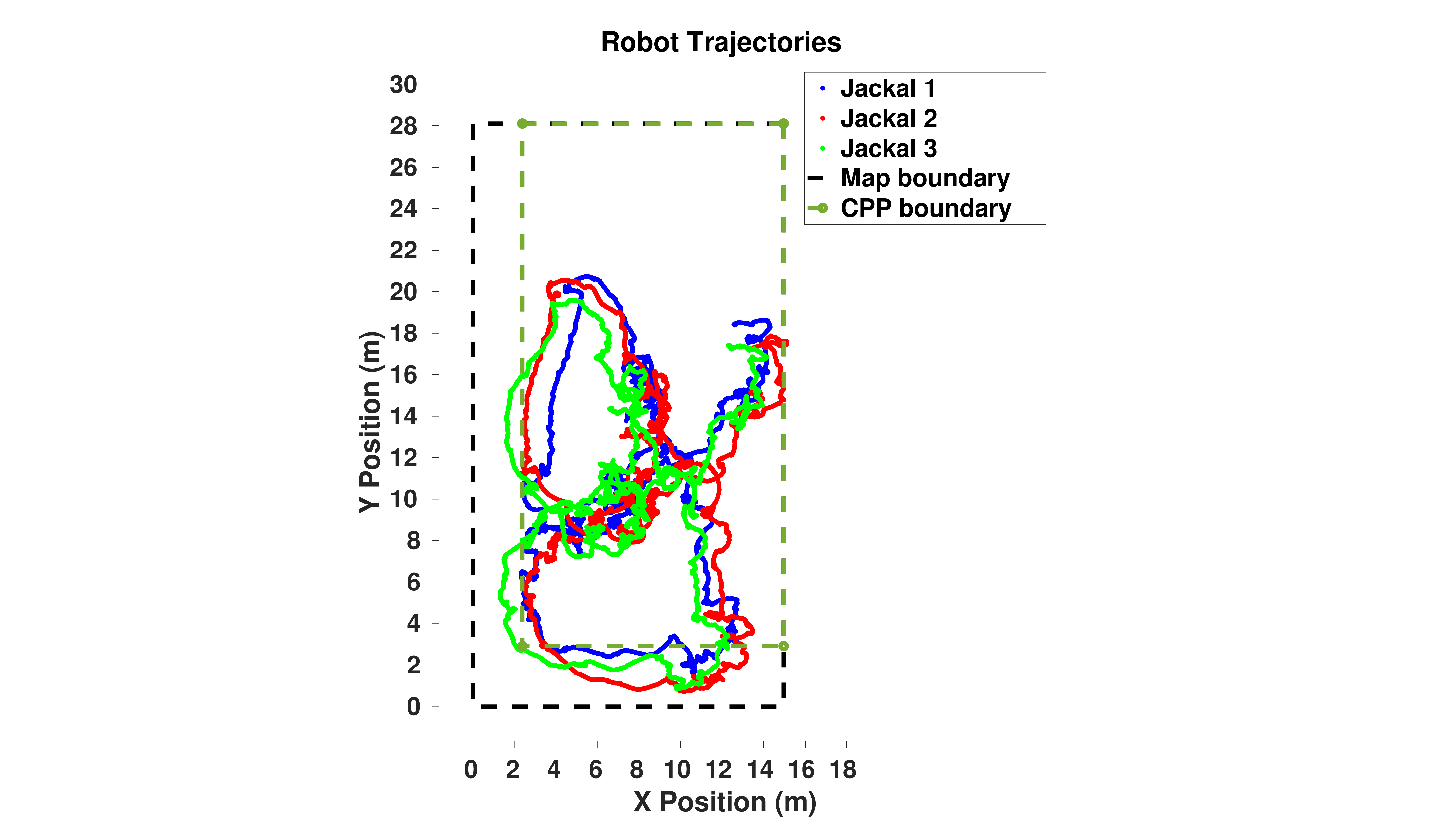} \\
			(b) \textit{AS.}\\[6pt]
		\end{tabular}
		\caption{\footnotesize Comparison of robot trajectories when guided by different algorithms: (a) CPP-AS and (b) AS.}
		\label{fig:robot_path}
	\end{center}
\end{figure}

\red{From Fig. \ref{fig:robot_path}, it is also interesting that the robot flock is often driven towards missed coverage areas or unplanned path areas (out of the CPP boundary) during the active sensing process. This means that the AS algorithm is able to select the next measurement location more accurately based on the PDF distribution provided by the CPP algorithm. In contrast, the AS algorithm, run alone, often selects the possible measurement positions that do not minimise the uncertainty of the gas field at the initial phase.}

Demonstration videos of all solutions compared in Experiment 2 can be viewed at the following address: AS: \url{https://tinyurl.com/23z2cjp7}; CPP-AS: \url{https://tinyurl.com/22cw27r4}. \red{As demonstrated in the CPP-AS method's video, our robot team effectively avoided obstacles using the closest safe angle-based obstacle avoidance. Notably, during the middle of the test, a human walked through the robots' path, twice entering the obstacle avoidance radius of Jackals 2 and 3. The UGVs promptly steered into the nearest safe area and resumed tracking their desired waypoints. Additionally, during the AS process, measurement positions are selected outside obstacle regions. This obstacle avoidance capability addresses a significant limitation of other coverage path planning or gas distribution mapping methods \cite{huang2020multi,jang2020multi,leong2022field,tran2023multi}.}

\subsubsection{Coverage Path Planning and Active Sensing with Real Source}

\red{Hardware-in-the-loop experimental results show that the CPP-AS method, integrated with a fixed one-way state-machine model, is highly effective for gas field estimation in dynamic urban terrains. To further validate these findings, we conducted real-time experiments using three mobile robots and a real radio source generated by a radio modem, testing each algorithm across five trials.} Given the unknown nature of the true field in practical scenarios, only ASPE, total exploration time, and the width of the highest concentration area (WHCA) metrics are employed for evaluation. Real-time experimental demonstrations can be viewed in the following videos: CPP-AS: \url{https://tinyurl.com/mr3h9vuj}, and AS \cite{leong2021estimation,leong2022field,tran2023multi}: \url{https://tinyurl.com/yckxum3r}.

\begin{figure}
	\centering
        \begin{subfigure}[b]{0.33\textwidth}
		\includegraphics[width=\textwidth]{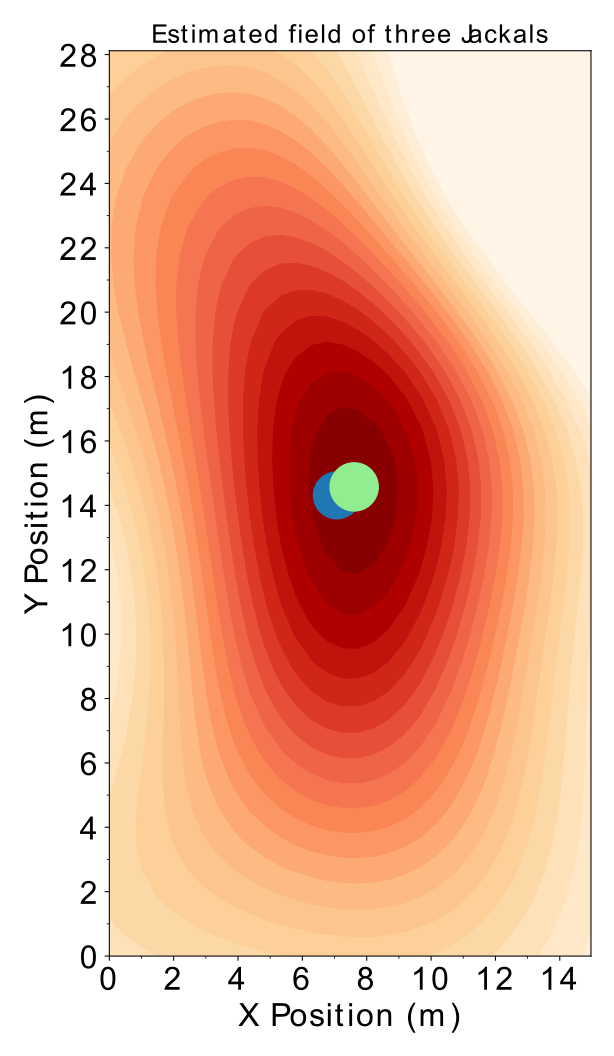}
		\caption{\textit{{CPP-AS.}}}
		\label{fig:pl3}
	\end{subfigure}
      \hfill
        \begin{subfigure}[b]{0.355\textwidth}
		\includegraphics[width=\textwidth]{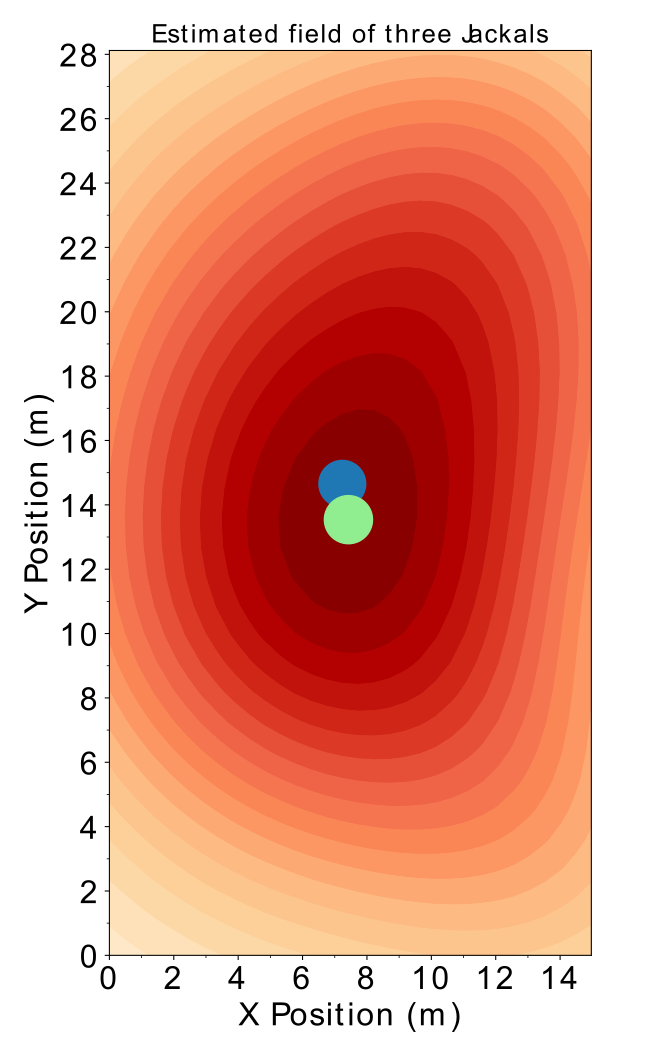}
		\caption{\textit{{AS.}}}
		\label{fig:ams3}
	\end{subfigure} 
		\caption{\footnotesize Local gas maps produced by the 3 Jackals after 243 measurements of (a) CPP-AS and 78 measurements of (b) AS \cite{leong2021estimation,leong2022field,tran2023multi} using the latest RSSI measurements. While the green circle represents the actual radio source, the blue circle indicates the estimated source location.}
		\label{fig:real_gas_maps}\vspace*{-1pt}
\end{figure}

As illustrated in Fig. \ref{fig:real_gas_maps}, \red{both methods accurately identify the single gas source in real-world experiments without detecting any false positive gas sources, consistent with conclusions drawn in hardware-in-the-loop results. The heat maps consistently reveal that the area with the highest gas concentration (defined by a maximum concentration value of 1.0) has a radius of approximately 1m, matching the predefined region with the highest gas concentration in the experimental setup. This observation underscores the robustness and accuracy of our CPP-AS approach.}

Furthermore, both CPP-AS and AS solutions exhibit higher accuracy in gas source localization compared to alternative strategies. \red{Although the mission time is not explicitly presented in the experimental results of existing literature, such as the partial differential equation model (PDEM)-based localization \cite{wiedemann2019model} (where three robots and a single simulated gas source are employed, with the source placed in the middle of a 3 m $\times$ 6 m lab environment, discretized into a 15 $\times$ 28 spatial sampling grid), and Geometrical Localization (GL) approaches \cite{guzey2022rf} (where a swarm of cooperative robots equipped with basic Received Signal Strength Indicator (RSSI) sensors coordinate their movements to localize a radio wave-emitting object),} the significance of this detail diminishes when considering the limitations revealed in the experimental outcomes of these methods. The PDEM-based approach exhibits reduced accuracy due to substantial mismatches between its assumptions and actual gas field conditions. After 140 measurements, it struggles to converge its gas model parameters to the real `ground truth' values. Similarly, the GL scheme yields a considerable ASPE of $52.66\pm2.11$m in noisy real-world experiments.

\begin{figure}
	\centering
	\includegraphics[width=27.5pc]{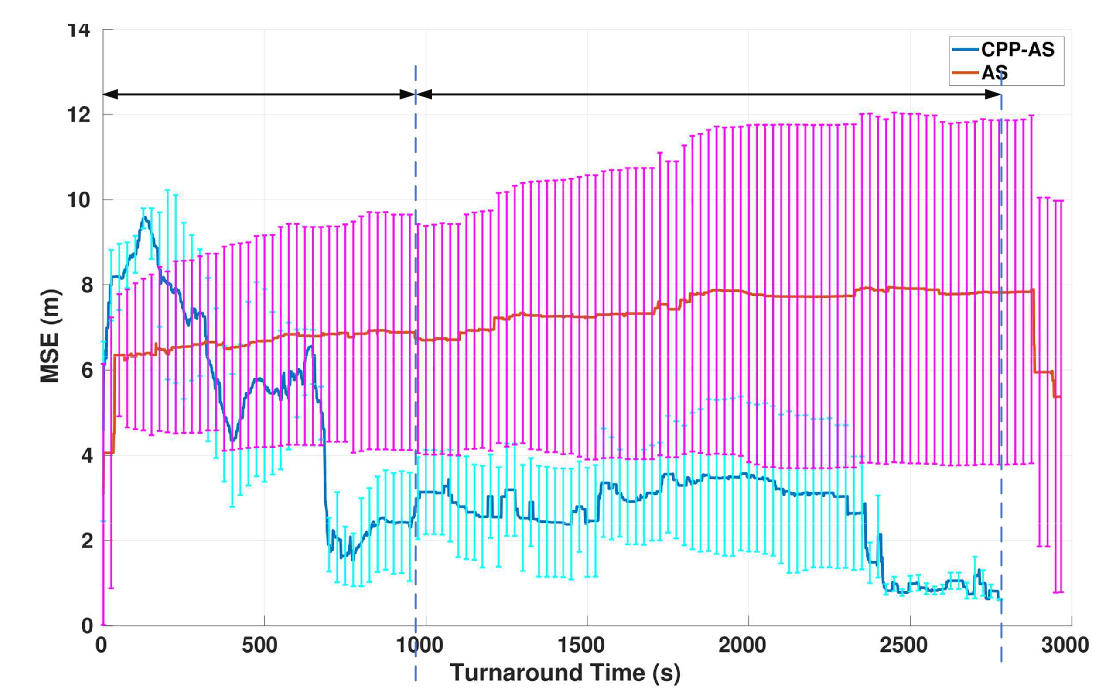}
	\caption{\footnotesize \red{Performance in real experiments: Mean square error of the estimated source location over time.}}
	\label{fig:mse_metrics_exp2}
\end{figure}

\red{In contrast, the AS approach, despite conducting only 78 measurements, obtains a significantly smaller ASPE of approximately $5.38\pm4.59$m after 2990s, as demonstrated in Fig. \ref{fig:mse_metrics_exp2}. Unlike the PDEM, where measurements concentrate close to the source locations, the AS algorithm selects points that minimize information-theoretic cost functions, as highlighted in the AS video. The swarm movement also significantly enhances the likelihood of measuring gas concentrations at neighboring points, thereby adeptly mitigating measurement noise and elevating overall accuracy compared to the PDEM and GL methods. Moreover, the AS approach strategically incorporates binary measurements to effectively tackle challenges associated with noisy raw gas data, presenting a noteworthy advantage over the GL approach.}

\red{The AS method, unfortunately, requires an additional 517s to attain the final ASPE level achieved by our CPP-AS method. This prolonged turnaround time and fewer measurements, in contrast to CPP-AS, stem from the incorrect selection of measurement locations during initialization, where a lack of prior information and high uncertainty in source parameters lead to significant increases in backtracking paths. Moreover, the AS approach's convergence time is not guaranteed due to the random selection of measurement points, as evident from the sizable 95\% confidence intervals. This poses a significant challenge, particularly in scenarios with limited battery life.}

\red{Compared to existing AS methods, our CPP-AS approach excels in achieving rapid and accurate source localization. As depicted in the CPP-AS video, the exploration phase begins with the multi-robot CPP and CSMC methods. Once the robots return to the starting point, the area coverage is completed, and an initial gas distribution map is generated with a reasonable error in gas source and field estimation. Subsequently, the active sensing phase is activated to enhance performance. The AS algorithm then selects the next measurement location more accurately based on the PDF distribution provided by the CPP algorithm. The exploration concludes after reaching the predefined maximum of 243 measurements, with the active sensing phase ending at precisely 2773s. The final estimated source location and its highest concentration field closely match the true source position and its highest concentration field, demonstrating the accuracy and effectiveness of our CPP-AS approach.}

The results depicted in Fig. \ref{fig:mse_metrics_exp2} demonstrate a significant convergence of the estimated parameters for the gas model within a remarkable duration of 2470 seconds. In all five trials, the CPP-AS method consistently and accurately identifies the radio source position by the end of the exploration period. \red{While the CPP-only approach results in a high APSE measure of $3.13\pm0.85$m, the CPP-AS method achieves a significantly lower ASPE of $0.61\pm0.11$m, which is approximately five times lower than the CPP-only approach. This indicates that many areas at the boundary of the true gas field were not visited or measured during the CPP phase, but were effectively covered during the AS phase.} Furthermore, the minimal 95\% confidence intervals of the ASPE during the final 303 seconds indicate steady and reliable convergence performance across multiple runs.


\red{These compelling outcomes are facilitated by our proposed state-machine model, effectively reducing exploration time. Moreover, our approach incorporates active sensing, allowing Jackal robots to intelligently navigate towards regions of high uncertainty and free from obstacles, thereby minimizing estimation errors. Through the formation of a swarm and the utilization of shared weights during the measurement of RSSI values at nearby points, we successfully mitigate the adverse effects of noise, further elevating the accuracy of our localization technique.}

In addition to source localisation and mapping, we address the critical problem of collision avoidance in a dynamic environment. In our video demonstration (\url{https://tinyurl.com/7hsj89fh}), \red{all robots successfully reached their goals without colliding with either unknown static obstacles, unknown dynamic obstacles, or other robots. Our proposed obstacle avoidance strategy, described in Section \ref{sub:FB}, enables a robot to automatically search for the closest safe area and steer towards it when the unknown obstacles travel within a pre-defined avoidance radius. Also, our active sensing technique never selects measurement points inside obstacles, significantly mitigating the risks of potential collisions. This is a significant advantage compared to conventional active sensing strategies, such as those in \cite{leong2021estimation,leong2022field, tran2023multi}.}


\subsubsection{\red{Impact of number of robots and obstacles on performance}}

\begin{figure}
	\centering
	\begin{subfigure}[b]{0.7\textwidth}
		\includegraphics[width=\textwidth]{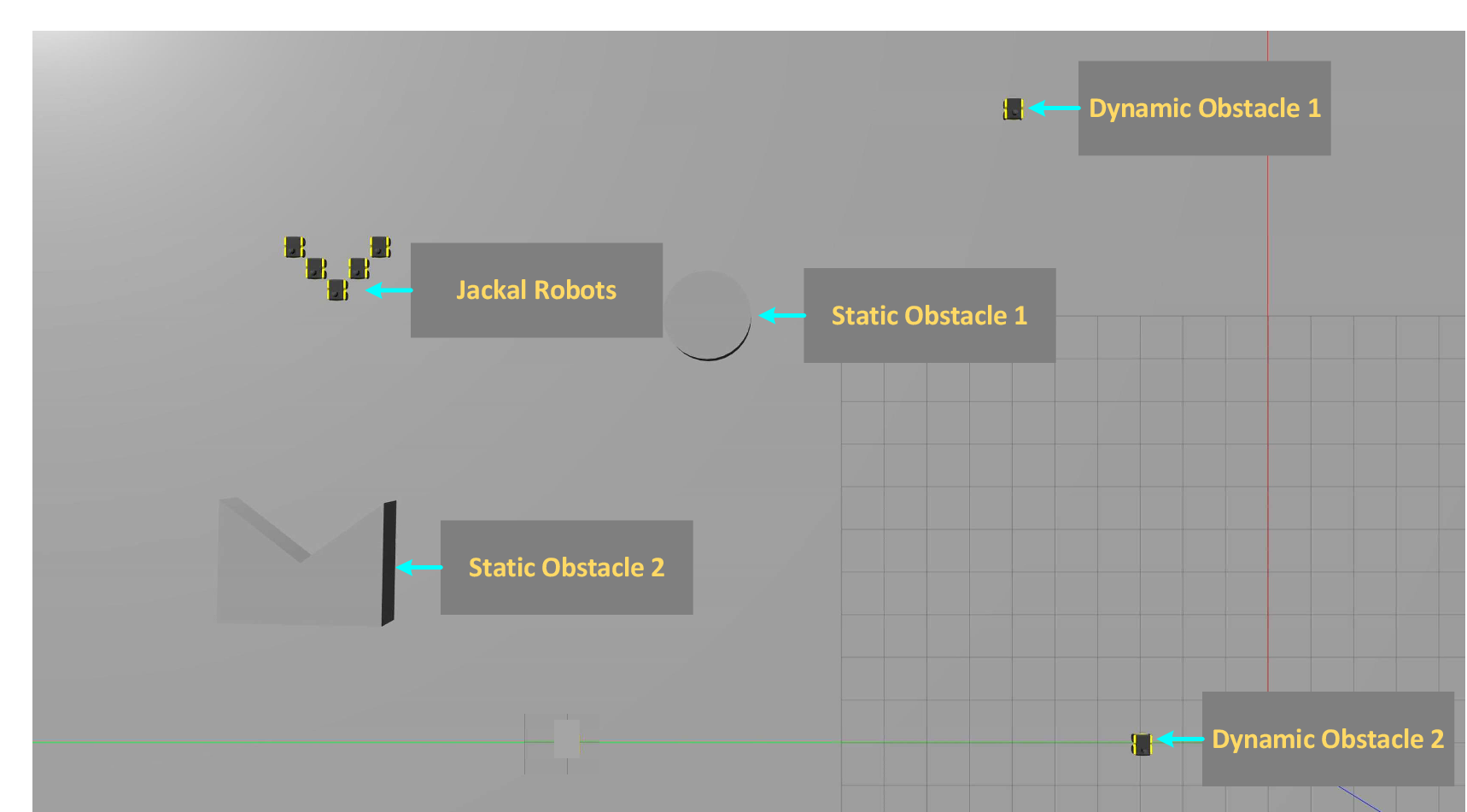}
		\caption{\textit{Scenario 1.}}
		\label{fig:urban}
	\end{subfigure}
	\begin{subfigure}[b]{0.7\textwidth}
		\includegraphics[width=\textwidth]{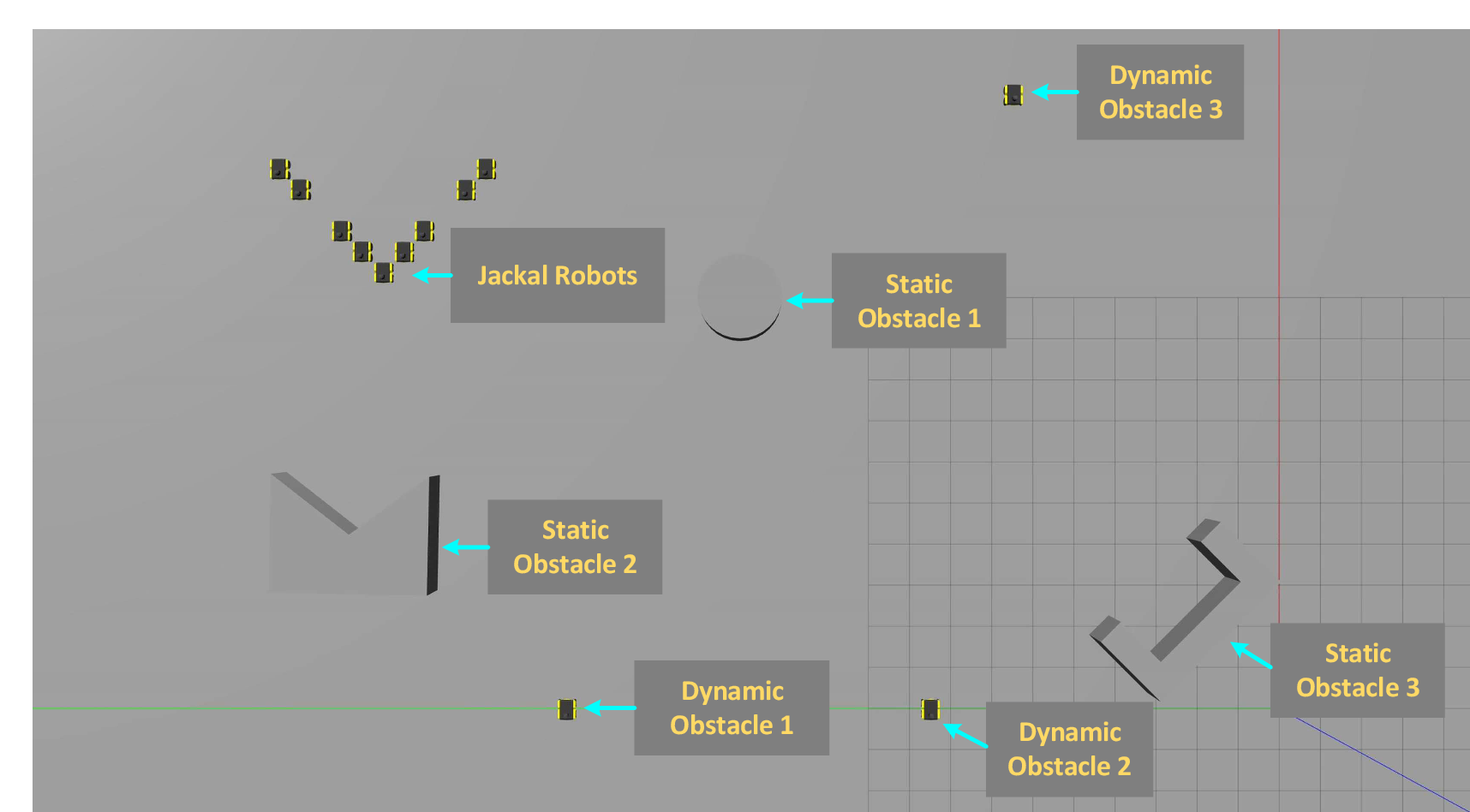}
		\caption{\textit{Scenario 2.}}
		\label{fig:map_sim}
	\end{subfigure}
	\caption{\footnotesize \red{Two complex urban environments.}}\label{fig:urban_ens}
\end{figure}

In our final experiment, we evaluate the impact of the proposed state-machine model on turnaround time, estimation accuracy (ANSME), and collision number by varying the number of robots and obstacles in the simulated two urban-like scenarios measuring 14.98m $\times$ 28.12m, as depicted in Fig. \ref{fig:urban_ens}. Scenario 1 involves five Jackal robots, two static obstacles \red{(e.g., a building and an office)}, and two dynamic obstacles, while Scenario 2 employs nine robots, three static obstacles \red{(e.g., two building blocks and an office)}, and three dynamic obstacles. Static obstacles feature intricate shapes, and dynamic obstacles are represented by Jackal robots undertaking additional tasks, navigating around the static obstacles. The true field, shown in Fig. \ref{fig:true_field}, remains consistent with the experiment in Section 4.3.2. Results are averaged over ten runs for each scenario.

\begin{figure}
	\begin{center}
		\begin{tabular}{cc}	
			\includegraphics[width=28pc]{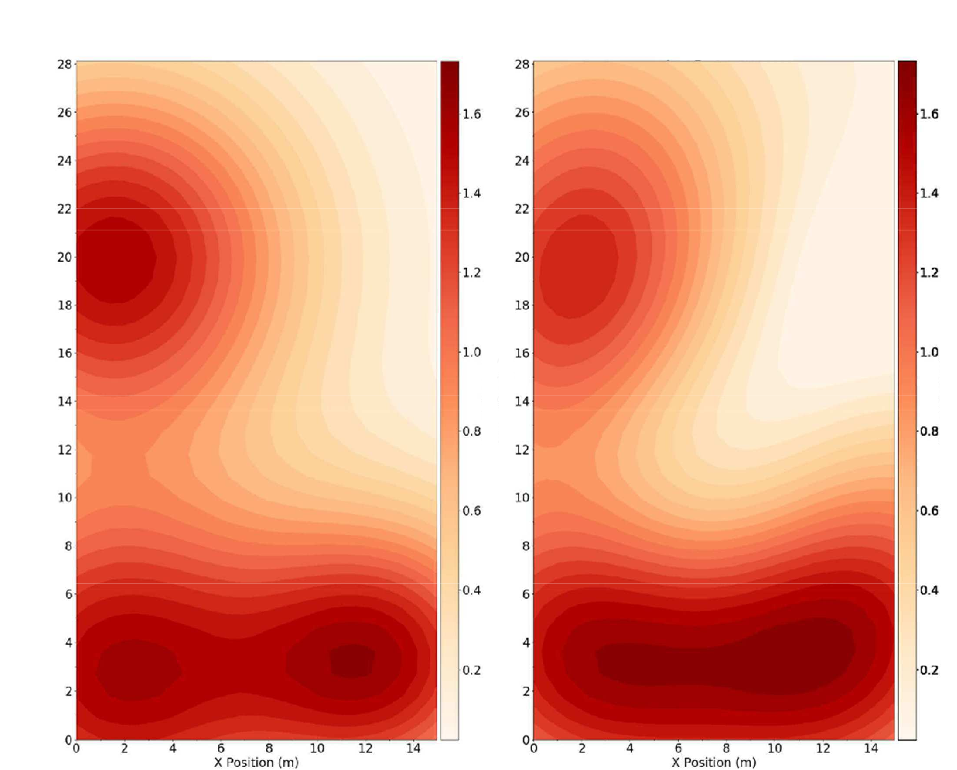} \\
			(a) \textit{Scenario 1.}\\[6pt]
			\includegraphics[width=28pc]{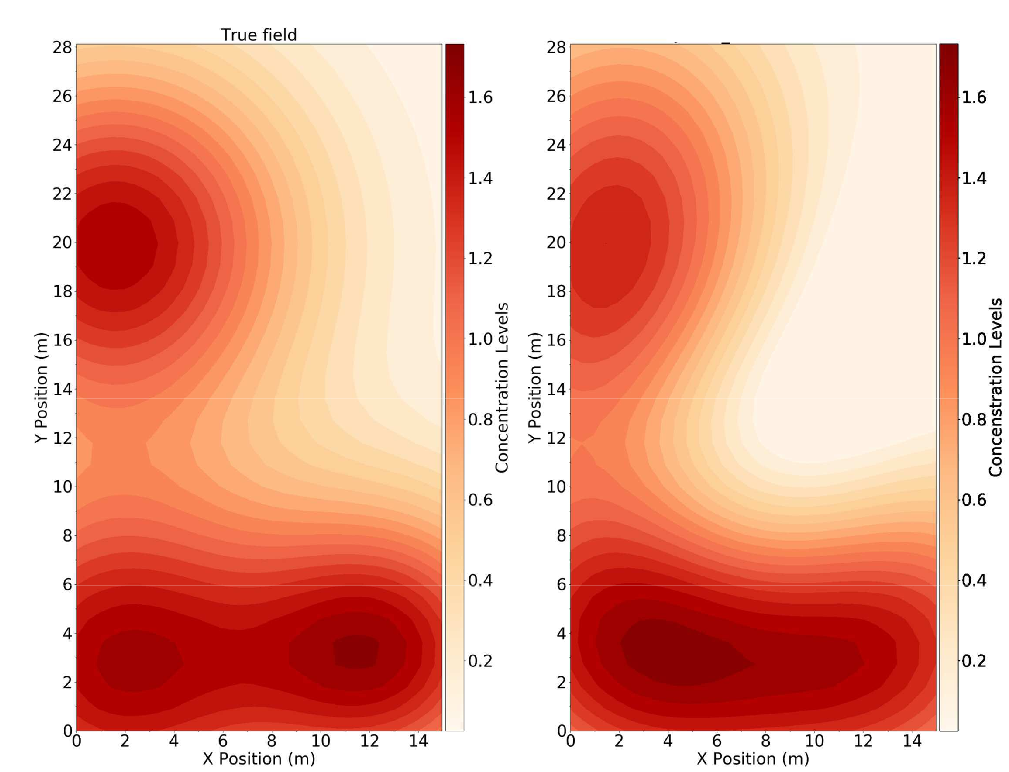} \\
			(b) \textit{Scenario 2.}\\[6pt]
		\end{tabular}
		\caption{\footnotesize \red{True field and two reconstructed fields obtained by (a) 5 Jackals in Scenario 1 and (b) 9 Jackals in Scenario 2, compared to the true gas field, after 300 measurements.}}
		\label{fig:gas_maps3}
	\end{center}
\end{figure}

\red{After completing the area coverage and returning to the initial point, the CPP approach generates a reasonable ANMSE of 0.082$\pm$0.027 for the 5-Jackal case and 0.099$\pm$0.047 for the 9-Jackal case.} Subsequently, active sensing is performed until 300 measurements are reached. Plots illustrating the approximations for two different trials, derived from parameter estimates obtained after 300 measurements of Algorithm \ref{alg:particle_sharing}, are presented in Fig. \ref{fig:gas_maps3}. These plots reveal gas fields remarkably close to the true field, particularly when the key features of the gas field, such as the number of gas sources, the concentration levels, the field shape, and the gas source locations, are achieved. Moreover, these fields align with those obtained in Fig. \ref{fig:gas_maps} of the virtual gas field experiment, indicating consistent results.

Table \ref{tab:deploy_time} provides compelling insights into the system's behavior. Firstly, as the number of robots engaged in active sensing increases, the turnaround time decreases, indicating improved efficiency. However, there appears to be evidence that the completion time may reduce slightly if an excessive number of robots is used. Notably, the 5-robot scenario takes only 355.11s more than the 9-robot case to obtain algorithm convergence in the same arena. This problem arises as robots navigate cautiously to avoid collisions, with increased density leading to more evasive actions and reduced focus on tracking the desired trajectory. Secondly, a nearly double increase in robots from 5 to 9 results in a steady improvement in gas mapping accuracy, decreasing the ANMSE figure from 0.076 $\pm$ 0.017 to 0.071 $\pm$ 0.010, attributed to a slower convergence rate of estimated parameters in the final phase. Importantly, even as the number of static and dynamic obstacles increases, Jackal robots, guided by our obstacle avoidance strategy, successfully complete the task without any collisions. This performance highlights the robustness and adaptability of the proposed method in dynamic environments, showcasing superior performance compared to the state-of-the-art active sensing methods \cite{wiedemann2019model,leong2022field, leong2022logistic, tran2023multi}.

\begin{table}[h]  
\caption{\footnotesize Effect of varying numbers of robots and obstacles on turnaround time, estimation accuracy, and collision number.}\label{tab:deploy_time}
\centering 
\resizebox{.97\textwidth}{!}{

\begin{tabular}{c c c c c c} 
\hline\hline   
Scenario &  No. of Robots and Obstacles & Turnaround Time (s)  & Field Estimation Accuracy & No. of Collisions
\\ [0.5ex]  
\hline   
1 & 5 Jackals and 4 Obstacles & 2995.950 $\pm$ 248.260 & 0.076 $\pm$ 0.017
 & 0.000 $\pm$ 0.000 \\ [1ex] 
2 & 9 Jackals and 6 Obstacles & 2640.840 $\pm$ 467.480 & 0.071 $\pm$ 0.010 & 0.000 $\pm$ 0.000 \\ [1ex] 

\hline 
\end{tabular}  }
\end{table}

\section{Conclusion and Future Work}

\subsection{Conclusion}
This paper has introduced a state-machine model for a multi-modal, multi-robot environmental sensing algorithm. This multi-modal algorithm integrates two different exploration algorithms: (1) coverage path planning using variable formations and (2) collaborative active sensing using multi-robot swarms. The state machine provides the logic for when to switch between these different sensing algorithms.

This paper also introduced a novel source localisation and mapping strategy for multiple robots. Consistent source concentration maps are constructed and updated over time by approximating and sharing the parameters of the Gaussian functions using the collaborative sequential Monte Carlo techniques, incorporating the novel exploration strategy to determine the measurement locations, enabling a reduced turnaround time for accurate field estimation.

Through our experimental comparisons, the CPP-AS algorithm outperforms the AS alone, as presented in \cite{leong2022field, leong2022logistic, tran2023multi}, by delivering superior map accuracy, faster convergence, and remarkable robustness in cluttered environments with multiple obstacles. Notably, the algorithm enables robots to exhibit flocking behavior and formation control, facilitating the sharing of particles and weights. Moreover, our proposed source localisation technique using binary measurements demonstrates significantly better performance compared to the existing localisation strategies that rely on analog measurements, such as the model-based and geometrical approaches \cite{wiedemann2019model,guzey2022rf}. Additionally, our approach reduces the impact of measurement noise by forming a swarm and measuring gas concentrations at both the target location and nearby points. This ensures more robust and reliable gas source localisation.

\subsection{Future Work}
For future work, there is potential to further expand the multi-model exploration model so that robots can utilise more of the pathways from Fig. \ref{fig:full_state}. These include pathways between CPP at different levels of resolution and pathways from AS back to CPP. 

We have also identified the possibility of developing an innovative active sensing method incorporating an expanded search space strategy. This approach will enable the robot swarm to gather more valuable information in a timely manner by dynamically adjusting the search space when the difference in probability density functions falls below a predefined threshold. 

\red{Despite conducting thorough simulations and real-time experiments with a simulated complex gas field and a radio source emulating a simple gas field, further validation in real-world scenarios remains essential. Future efforts will prioritize testing the CPP-AS algorithm in environments with actual gas field leaks. These real-world deployments are challenging due to the unpredictable nature of CBRN agents, complex experimental setups, strict safety standards, and rigorous regulatory requirements, which necessitate meticulous planning and execution. Claims about real-world performance will be framed as hypotheses for future testing, acknowledging the need to demonstrate effectiveness under actual operational conditions. Additionally, we will enhance the clarity of current claims regarding performance within the controlled scope of our experiments to provide a comprehensive understanding of the algorithm's capabilities. Such real-world tests will provide invaluable insights into the algorithm's effectiveness and reliability in dynamic and unpredictable settings, thereby robustly enhancing its practical applicability.}

\section*{Acknowledgment}
This work was supported by the Australian Defence Science \& Technology  Group (DSTG) under a grant titled ``Multi-Robot Systems for Operation in CBRN Environments'' with agreement number ID10096.

\bibliographystyle{apalike}
\bibliography{Radio_gas_estimation}


\end{document}